\documentclass[acmsmall,screen,authorversion,nonacm]{acmart}

\usepackage{hyperref}
\usepackage{url}

\usepackage{booktabs}       
\usepackage{amsfonts}       
\usepackage{microtype}      
\usepackage{xcolor}         

\usepackage{natbib}
\usepackage{graphicx}
\usepackage{subfigure}
\usepackage{amsmath}
\usepackage{amsthm}
\usepackage{multirow}
\usepackage{algorithm}
\usepackage{algorithmic}

\newtheorem{theorem}{Theorem}
\newtheorem{definition}{Definition}
\newtheorem*{remark}{Remark}
\newtheorem{proposition}{Proposition}
\makeatletter
\newsavebox{\@brx}
\newcommand{\llangle}[1][]{\savebox{\@brx}{\(\m@th{#1\langle}\)}%
  \mathopen{\copy\@brx\kern-0.5\wd\@brx\usebox{\@brx}}}
\newcommand{\rrangle}[1][]{\savebox{\@brx}{\(\m@th{#1\rangle}\)}%
  \mathclose{\copy\@brx\kern-0.5\wd\@brx\usebox{\@brx}}}

\makeatother
\AtBeginDocument{%
  \providecommand\BibTeX{{%
    \normalfont B\kern-0.5em{\scshape i\kern-0.25em b}\kern-0.8em\TeX}}}

\setcopyright{acmcopyright}
\copyrightyear{2018}
\acmYear{2018}
\acmDOI{XXXXXXX.XXXXXXX}

\acmConference[Conference acronym 'XX]{Make sure to enter the correct
  conference title from your rights confirmation emai}{June 03--05,
  2018}{Woodstock, NY}
\acmPrice{15.00}
\acmISBN{978-1-4503-XXXX-X/18/06}

\begin{document}

\title{Unit Ball Model for Embedding Hierarchical Structures in the Complex Hyperbolic Space}

\author{Huiru Xiao}
\email{hxiaoaf@cse.ust.hk}
\affiliation{%
  \institution{Hong Kong University of Science and Technology}
  \streetaddress{Clear Water Bay}
  \city{Kowloon}
  \country{Hong Kong SAR, China}
}

\author{Caigao Jiang}
\affiliation{%
  \institution{Ant Group}
  \streetaddress{No.556, Xixi Road, Xihu}
  \city{Hangzhou}
  \state{Zhejiang}
  \country{China}}
\email{caigao.jcg@antgroup.com}

\author{Yangqiu Song}
\affiliation{%
  \institution{Hong Kong University of Science and Technology}
  \streetaddress{Clear Water Bay}
  \city{Kowloon}
  \country{Hong Kong SAR, China}}
\email{yqsong@cse.ust.hk}

\author{James Zhang}
\affiliation{%
  \institution{Ant Group}
  \streetaddress{No.556, Xixi Road, Xihu}
  \city{Hangzhou}
  \state{Zhejiang}
  \country{China}
}
\email{james.z@antgroup.com}

\author{Junwu Xiong}
\affiliation{%
 \institution{Ant Group}
\streetaddress{No.556, Xixi Road, Xihu}
  \city{Hangzhou}
  \state{Zhejiang}
  \country{China}}
 \email{junwu.xjw@antgroup.com}


\begin{abstract}
  Learning the representation of data with hierarchical structures in the hyperbolic space attracts increasing attention in recent years. 
Due to the constant negative curvature, the hyperbolic space resembles tree metrics and captures the tree-like properties naturally, which enables the hyperbolic embeddings to improve over traditional Euclidean models.
However, many real-world hierarchically structured data such as taxonomies and multitree networks have varying local structures and they are not trees, thus they do not ubiquitously match the constant curvature property of the hyperbolic space. To address this limitation of hyperbolic embeddings, we explore the complex hyperbolic space, which has the variable negative curvature, for representation learning. Specifically, we propose to learn the embeddings of hierarchically structured data in the unit ball model of the complex hyperbolic space. The unit ball model based embeddings have a more powerful representation capacity to capture a variety of hierarchical structures. Through experiments on synthetic and real-world data, we show that our approach improves over the hyperbolic embedding models significantly. We also explore the competence of complex hyperbolic geometry on the multitree structure and $1$-$N$ structure. Our codes are available at \url{https://github.com/HKUST-KnowComp/UnitBall}.
\end{abstract}




\maketitle

\section{Introduction}
\label{sec:introduction}
Representation learning of data with hierarchical structures is an {important machine learning task} with many applications, such as taxonomy induction~\citep{DBLP:conf/acl/FuGQCWL14} and hypernymy detection~\citep{DBLP:conf/acl/ShwartzGD16}. 
In recent years, the hyperbolic embeddings~\citep{DBLP:conf/nips/NickelK17,DBLP:conf/icml/NickelK18} have been proposed to improve the traditional Euclidean embedding models.
The constant negative curvature of the hyperbolic space produces a manifestation that the hyperbolic space can be regarded as a continuous approximation to trees~\citep{PhysRevE.82.036106}. The hyperbolic space is capable of embedding any finite tree while preserving the distances approximately~\citep{gromov1987hyperbolic}. 

However, most {real-world} hierarchical data do not belong to tree structures since they can have varying local structures while being tree-like globally. For example, the taxonomies such as WordNet~\citep{DBLP:journals/cacm/Miller95} and YAGO~\citep{DBLP:conf/www/SuchanekKW07} contain many $1$-$N$ ($1$ child links to multiple parents) cases and multitree structures~\citep{DBLP:journals/jct/GriggsLL12}, which are much more complicated than the tree structure.
{In consequence, the hyperbolic space which resembles tree metrics has limitations on capturing the general {hierarchically structured data}.}

To address the challenge, 
in this paper, we propose a new approach to learning the embeddings of {hierarchically structured data}. Specifically, we embed the hierarchical data into the unit ball model of the complex hyperbolic space. The unit ball model is a projective geometry based model to identify the complex hyperbolic space. One of the main differences between the complex and the real hyperbolic space is that the curvature is non-constant in the complex hyperbolic space. In practice, the variable negative curvature makes the complex hyperbolic space more flexible in handling varying structures while the tree-like properties are still retained.

For empirical evaluation, we evaluate different geometrical embedding models on various {hierarchically structured data}, including synthetic graphs and real-world data. The experimental results demonstrate the advantages of our approach. In addition, we investigate two specific structures in which complex hyperbolic geometry shows outstanding performances, namely the multitree structure and $1$-$N$ structure, which are highly common and typical in real-world data.

To summarize, our work has the following main contributions:
\begin{enumerate}
\item We present a novel embedding approach based on the complex hyperbolic geometry to handle data with complicated and various hierarchical structures. To the best of our knowledge, our work is the first to propose complex hyperbolic embeddings.
\item We introduce the embedding algorithm in the unit ball model of the complex hyperbolic space. We formulate the learning and Riemannian optimization in the unit ball model.
\item We evaluate our approach with experiments on an extensive range of synthetic and real-world data and show the remarkable improvements of our approach. 
\end{enumerate}



\section{Related Work}
\label{sec:related work}
\paragraph{Hyperbolic embeddings.} Hyperbolic embedding methods have become the leading approach for representation learning of hierarchical structures. \citet{DBLP:conf/nips/NickelK17} learned the representations of hierarchical graphs in the Poincar\'e ball model of the hyperbolic space and outperformed the Euclidean embedding methods for taxonomies. The Poincar\'e embedding model was then improved by follow-up works on hyperbolic emebddings~\citep{DBLP:conf/icml/GaneaBH18,DBLP:conf/icml/NickelK18}.
These methods learned the hyperbolic embeddings by Riemannian optimization~\citep{DBLP:journals/tac/Bonnabel13}, which was further improved by the Riemannian adaptive optimization~\citep{DBLP:conf/iclr/BecigneulG19}. 

Another branch of study~\citep{DBLP:conf/icml/SalaSGR18,DBLP:conf/nips/SonthaliaG20} learned the hyperbolic embeddings through combinatorial construction.  
Instead of optimizing the soft-ranking loss by Riemannian optimization as in \citep{DBLP:conf/nips/NickelK17,DBLP:conf/icml/NickelK18}, the construction-based methods minimize the reconstruction distortion by combinatorial construction. 
However, both the optimization-based and construction-based hyperbolic embeddings suffer from the limitation in hierarchical graphs with varying local structures. To tackle the challenge, \citet{DBLP:conf/iclr/GuSGR19} extended the construction-based method by jointly learning the curvature and the embeddings of data in a product manifold. Although it can provide a better representation than a single space with constant curvature,
it is impractical to search for the best manifold combination among enormous combinations for each new structure.

Motivated by the promising results of previous works, extensions to the multi-relational graph hyperbolic embeddings~\citep{DBLP:conf/nips/BalazevicAH19,DBLP:conf/acl/ChamiWJSRR20,DBLP:conf/emnlp/SunCHWDZ20} and hyperbolic neural networks~\citep{DBLP:conf/nips/GaneaBH18,DBLP:conf/iclr/GulcehreDMRPHBB19,DBLP:conf/nips/LiuNK19,DBLP:conf/nips/ChamiYRL19,DBLP:conf/nips/ZhuP00C020,DBLP:journals/corr/abs-2104-06942,DBLP:journals/corr/abs-2006-08210} were explored. 
Notably, ~\citep{DBLP:conf/nips/ChamiYRL19,DBLP:conf/acl/ChamiWJSRR20} leverages the trainable curvature to compensate for the disparity between the actual data structures and the constant-curvature hyperbolic space, where each layer in the graph neural network or each relation in the multi-relational graph has its own curvature parameterization.
The hyperbolic learning also inspired other research tasks and applications, such as classification~\citep{DBLP:conf/aistats/ChoD0B19}, image reconstruction~\citep{DBLP:conf/iclr/SkopekGB20}, text generation~\citep{DBLP:conf/naacl/DaiGCTCL21}, etc.

\paragraph{Complex embeddings.} The traditional knowledge graph embeddings were learned in the real Euclidean space~\citep{DBLP:conf/icml/NickelTK11,DBLP:conf/nips/BordesUGWY13,DBLP:journals/corr/YangYHGD14a} and were used for knowledge graph inference and reasoning. In recent years, several works suggested utilizing the complex Euclidean space for inferring more relation patterns, such as ComplEx~\citep{DBLP:conf/icml/TrouillonWRGB16} and RotatE~\citep{DBLP:conf/iclr/SunDNT19}. These models have been demonstrated to be effective in knowledge graph embeddings. The success of the complex embeddings reveals the potential of the complex space and inspires us to explore the complex hyperbolic space.

\section{Preliminaries}
\label{sec:preliminaries}
\subsection{Basic Definitions}
Before introducing the hyperbolic geometry and the complex hyperbolic geometry, we need to explain and define some related concepts. The first one is \textit{curvature}, which describes the curve of Riemannian manifolds and controls the rate of geodesic deviation.\footnote{In this paper, \textit{curvature} refers to the \textit{sectional curvature.}}
\subsubsection{Curvature}
\label{sec:curvature}

\begin{definition}[Curvature]
Given a Riemannian manifold and two linearly independent tangent vectors at the same point, $\mathbf{u}$ and $\mathbf{v}$, the \textbf{(sectional) curvature} is defined as
\begin{equation*}
    K(\mathbf{u},\mathbf{v})=\frac{\langle R(\mathbf{u},\mathbf{v})\mathbf{v},\mathbf{u}\rangle}{\langle \mathbf{u},\mathbf{u}\rangle\langle\mathbf{v},\mathbf{v}\rangle-\langle \mathbf{u},\mathbf{v}\rangle^2},
\end{equation*}
where $R$ is the Riemann curvature tensor, defined by the convention $R(\mathbf{u},\mathbf{v})\mathbf{w}=\nabla_\mathbf{u}\nabla_\mathbf{v}\mathbf{w}-\nabla_\mathbf{v}\nabla_\mathbf{u}\mathbf{w}-\nabla_{[\mathbf{u},\mathbf{v}]}\mathbf{w}$. Here $\nabla$ indicates the Levi-Civita connection, whose definitions are given below.
\end{definition}
Before defining Levi-Civita connection, we need to first define the affine connection.
\begin{definition}[Affine connection]
Let $M$ be a smooth manifold and let $\Gamma(\mathcal{T}M)$ be the space of vector fields on $M$, i.e., the space of smooth sections of the tangent bundle $\mathcal{T}M$. Then an \textbf{affine connection} on M is a bilinear map
\begin{align*}
    \Gamma(\mathcal{T}M)\times\Gamma(\mathcal{T}M)&\to\Gamma(\mathcal{T}M) \\
    (X,Y)&\mapsto\nabla_XY,
\end{align*}
such that for all $f$ in the set of smooth functions on $M$, written $C^\infty(M,\mathbb{R})$, and all vector fields $X, Y$ on $M$:
\begin{enumerate}
    \item $\nabla_{fX}Y=f\nabla_XY$, i.e., $\nabla$ is $C^\infty(M,\mathbb{R})$-linear in the first variable;
    \item $\nabla_X(fY)=\partial_XfY+f\nabla_XY$, where $\partial_X$ denotes the directional derivative, i.e., $\nabla$ satisfies \textit{Leibniz rule} in the second variable.
\end{enumerate}
\end{definition}

Next, we define the Levi-Civita connection.
\begin{definition}[Levi-Civita connection]
An affine connection $\nabla$ is called a \textbf{Levi-Civita} connection if
\begin{enumerate}
    \item it preserves the metric, i.e., $\nabla g=0$.
    \item it is torsion-free, i.e., for any vector fields $X$ and $Y$ we have $\nabla_XY-\nabla_YX=[X,Y]$, where $[X,Y]$ is the Lie bracket of the vector fields $X$ and $Y$.
\end{enumerate}
\end{definition}

\subsubsection{$\delta$-Hyperbolicity}
\label{sec:delta-hyperbolicity}
 Here we give the definition of $\delta$-hyperbolicity~\citep{gromov1987hyperbolic}, which measures the tree-likeness of graphs. The lower $\delta$ corresponds to the more tree-like graph. Trees have $0$ $\delta$-hyperbolicity.
\begin{definition}[$\delta$-hyperbolicity]
\label{def:hyperbolicity}
Let $a,b,c,d$ be vertices of the graph $G$. Let $S_1$, $S_2$ and $S_3$ be
\begin{equation*}
    S_1=dist(a,b)+dist(d,c), 
    S_2=dist(a,c)+dist(b,d), 
    S_3=dist(a,d)+dist(b,c).
\end{equation*}
Suppose $M_1$ and $M_2$ are the two largest values among $S_1$, $S_2$, $S_3$ and $M_1\geq M_2$. Define ${hyp}(a,b,c,d)=M_1-M_2$. Then the \textbf{$\delta$-hyperbolicity} of $G$ is defined as
\begin{equation*}
    \delta(G)=\frac{1}{2}\max_{a,b,c,d\in V(G)}{hyp}(a,b,c,d).
\end{equation*}
That is, $\delta(G)$ is the maximum of $hyp$ over all possible $4$-tuples $(a,b,c,d)$ divided by $2$.
\end{definition}

\subsection{Hyperbolic Geometry}
Hyperbolic space\footnote{In this paper, we use \textit{hyperbolic space} to refer to real hyperbolic space and \textit{hyperbolic embeddings} to refer to real hyperbolic embeddings for avoiding wordiness.} is a homogeneous space with constant negative curvature.
In the hyperbolic space $\mathbb{H}_\mathbb{R}^n(K)$ of dimension $n$ and curvature $K$, the volume of a ball grows exponentially with its radius $\rho$:
\begin{equation}
    vol(B_{\mathbb{H}_\mathbb{R}^n(K)}(\rho))\sim e^{\sqrt{-K}(n-1)\rho}.
\end{equation}

Contrastively, in the Euclidean space $\mathbb{E}^n$, the curvature is $0$ and the volume of a ball grows polynomially with its radius:
\begin{equation}
    vol(B_{\mathbb{E}^n}(\rho))=\frac{\pi^{n/2}}{\Gamma(n/2)}\rho^n\sim\rho^n.
\end{equation}

The exponential volume growth rate enables the hyperbolic space to have powerful representation capability for tree structures since the number of nodes grows exponentially with the depth in a tree, while the Euclidean space is too flat and narrow to embed trees.

\subsection{Complex Hyperbolic Geometry}
\label{sec:complex hyperbolic geometry}
Complex hyperbolic space is a homogeneous space of variable negative curvature. Its ambient Hermitian vector space $\mathbb{C}^{n,1}$ is the complex Euclidean space $\mathbb{C}^{n+1}$ endowed with some Hermitian form $\llangle\mathbf{z},\mathbf{w}\rrangle$, where $\mathbf{z},\mathbf{w}\in\mathbb{C}^{n+1}$. Then the Hermitian space $\mathbb{C}^{n,1}$ can be divided into three subsets: $V_-=\{\mathbf{z}\in\mathbb{C}^{n,1}|\llangle\mathbf{z},\mathbf{z}\rrangle<0\}$, $V_0=\{\mathbf{z}\in\mathbb{C}^{n,1}-\{\mathbf{0}\}|\llangle\mathbf{z},\mathbf{z}\rrangle=0\}$, and $V_+=\{\mathbf{z}\in\mathbb{C}^{n,1}|\llangle\mathbf{z},\mathbf{z}\rrangle>0\}$. Let $\mathbb{P}$ be a projection map $\mathbb{P}: \mathbb{C}^{n,1}-\{z_{n+1}=0\}\to\mathbb{C}^n$, i.e.,
\begin{equation}
\label{eq:projection map}
    \mathbb{P}: 
    \begin{bmatrix}
z_1\\
\vdots\\
z_{n+1}
\end{bmatrix}\mapsto\begin{bmatrix}
z_1/z_{n+1}\\
\vdots \\
z_n/z_{n+1}
\end{bmatrix},
\text{where } z_{n+1}\neq0.
\end{equation}

Then the complex hyperbolic space $\mathbb{H}_\mathbb{C}^n$ and its boundary $\partial\mathbb{H}_\mathbb{C}^n$ are defined using the projectivization:
\begin{equation}
\label{eq:complex hyperbolic space}
    \mathbb{H}_\mathbb{C}^n=\mathbb{P}V_-, \qquad \partial\mathbb{H}_\mathbb{C}^n=\mathbb{P}V_0.
\end{equation}

\subsubsection{The Unit Ball Model}
\label{sec:unit ball model}
The unit ball model is one model used to identify the complex hyperbolic space, which can be derived via the projective geometry~\citep{goldman1999complex}. We now provide the necessary derivation.

Take the abovementioned Hermitian form of $\mathbb{C}^{n,1}$ to be a standard Hermitian form:
\begin{equation}
\label{eq:Hermitian form}
    \llangle\mathbf{z},\mathbf{w}\rrangle=z_1\overline{w_1}+\cdots+z_n\overline{w_n}-z_{n+1}\overline{w_{n+1}},
\end{equation}
where $\overline{w}$ is the conjugate of $w$.
Take $z_{n+1}=1$ in the projection map $\mathbb{P}$ in Eq. (\ref{eq:projection map}). Then from Eq. (\ref{eq:complex hyperbolic space}), we can derive the formula of the unit ball model:
\begin{equation}
\label{eq:unit ball}
    \mathcal{B}_\mathbb{C}^n=\mathbb{P}(\{\mathbf{z}\in\mathbb{C}^{n,1}|\llangle\mathbf{z},\mathbf{z}\rrangle<0\})=\{(z_1,\cdots,z_n,1)||z_1|^2+\cdots+|z_n|^2<1\}.
\end{equation}

The metric on $\mathcal{B}_\mathbb{C}^n$ is Bergman metric, which takes the formula below in $2$-d case:
\begin{equation}
\label{eq:metric tensor}
    ds^2=\frac{-4}{\llangle\mathbf{z},\mathbf{z}\rrangle^2}\det\begin{bmatrix}
\llangle\mathbf{z},\mathbf{z}\rrangle & \llangle d\mathbf{z},\mathbf{z}\rrangle\\
\llangle\mathbf{z},d\mathbf{z}\rrangle & \llangle d\mathbf{z},d\mathbf{z}\rrangle
\end{bmatrix}.
\end{equation}

The distance function on $\mathcal{B}_\mathbb{C}^n$ is given by
\begin{equation}
\label{eq:distance}
    d_{\mathcal{B}_\mathbb{C}^n}(\mathbf{z},\mathbf{w})=arcosh\Big(2\frac{\llangle\mathbf{z},\mathbf{w}\rrangle\llangle\mathbf{w},\mathbf{z}\rrangle}{\llangle\mathbf{z},\mathbf{z}\rrangle\llangle\mathbf{w},\mathbf{w}\rrangle}-1\Big),
\end{equation}

The distance function maintains the tree-like metric properties. When the points are very close to the origin, it approximates to the Euclidean distance. Additionally, when a point is closer to the origin, it has relatively smaller distances to the other points. Correspondingly, the points near the boundary have very large distances from each other. Therefore, in ideal conditions, the root node of a tree is embedded in the origin while the deeper nodes are embedded farther away from the origin. Recall that the distance function in the real hyperbolic space~\citep{DBLP:conf/nips/NickelK17} has similar properties since the real hyperbolic space, as the totally geodesic subspace of the complex hyperbolic space, inherits the tree-like metrics. More details about the Bergman metric and distance function can be referred to Chapter 3.1 in~\citep{goldman1999complex}.

Note that there are other choices of the Hermitian form $\llangle\mathbf{z},\mathbf{w}\rrangle$, which corresponds to other models of complex hyperbolic geometry, such as the Siegel domain model. We choose the unit ball model for the relatively simple formula as well as convenient computations of the metric and distance function.

\subsubsection{Variable Negative Curvature}
\label{sec:curvature proof}
The curvature of the complex hyperbolic space is summarized by~\citep{goldman1999complex} as follows:
\begin{theorem}
\label{theo:complex hyperbolic curvature}
The curvature is not constant in $\mathbb{H}_\mathbb{C}^n$. It is pinched between $-1$ (in the directions of complex projective lines) and $-1/4$ (in the directions of totally real planes).
\end{theorem}

Before proving Theorem 1, we need to introduce the definition of \textit{K\"{a}hler structure}~\citep{mok1989metric}.
\begin{definition}[K\"{a}hler structure]
\label{def:Kahler}
A \textbf{K\"{a}hler structure} can be defined in any of the following equivalent ways:
\begin{enumerate}
    \item A complex structure with a closed, positive $(1,1)$-form.
    \item A Riemannian structure with a complex structure such that the corresponding exterior $2$-form is closed.
    \item A symplectic structure with a compatible integrable almost complex structure which is positive.
\end{enumerate}
\end{definition}

Recall that the complex hyperbolic space $\mathbb{H}_\mathbb{C}^n$ is defined using the projectivization of the negative zone with a Hermitian form $\llangle\mathbf{z},\mathbf{w}\rrangle$. Denote $\omega$ as the imaginary part of the Hermitian form $\llangle,\rrangle$, i.e., $\omega(\mathbf{z},\mathbf{w})=\frac{1}{2i}(\llangle\mathbf{z},\mathbf{w}\rrangle-\llangle\mathbf{w},\mathbf{z}\rrangle)$, then according to~\citep{goldman1999complex}, the metric $\omega$ is \textit{positive} and \textit{closed}, and necesssarily has type $(1,1)$. Then by the first definition in Definition \ref{def:Kahler}, $\mathbb{H}_\mathbb{C}^n$ is a K\"{a}hler structure.

Let $M$ be a K\"{a}hler manifold and $\mathbf{z}\in M$. Denote $\mathcal{T}_{\mathbf{z}}M$ as the tangent space of $M$ at $\mathbf{z}$ and $J:\mathcal{T}M\to\mathcal{T}M$ is an endomorphism. As proved in~\citep{kobayashi1963foundations}, the curvature of real $2$-planes in the tangent space $\mathcal{T}_{\mathbf{z}}M$ has the following properties:
\begin{theorem}
\label{theorem:constant holomorphic curvature}
Let $M$ be a connected K\"{a}hler manifold of complex dimension $n\geq2$. If the holomorphic sectional curvature $K(p)$, where $p$ is a plane in $\mathcal{T}_{\mathbf{z}}M$ invariant by $J$, depends only on $\mathbf{z}$, then $M$ is a space of constant holomorphic sectional curvature.
\end{theorem}

Next, we give a proposition in~\citep{kobayashi1963foundations}, which is about the curvature of a plane.
\begin{proposition}
\label{prop:plane curvature}
If $\mathbf{u},\mathbf{v}$ is an orthonormal basis for a plane $p$ and if we set the curvature of $p$ as $K(p)=R(\mathbf{u},\mathbf{v})$, where $R(\mathbf{u},\mathbf{v})$ is the Riemann curvature tensor, then
\begin{equation*}
    K(p)=\frac{1}{4}(1+3\cos^2\alpha(p)),
\end{equation*}
where $\alpha(p)$ is the angle between $p$ and $J(p)$.
\end{proposition}

Finally, we prove Theorem 1 as follows.
\begin{proof}
Let $M$ be a K\"{a}hler manifold and $\mathbf{z}\in M$. From Theorem \ref{theorem:constant holomorphic curvature}, the corresponding sectional curvature function of real $2$-planes in $\mathcal{T}_{\mathbf{z}}M$ is completely determined by the sectional curvature function restricted to complex lines in $\mathcal{T}_{\mathbf{z}}M$. If the sectional curvature of every complex line in $\mathcal{T}M$ equals $\kappa$, then $M$ is said to have constant holomorphic sectional curvature $\kappa$.

Then from Proposition \ref{prop:plane curvature}, we can know that in this case, the sectional curvature of a $2$-dimensional subspace $S\subset\mathcal{T}M$ is
\begin{equation}
\label{eq:K(S)}
    K(S)=\kappa\frac{1+3\cos^2\alpha(S)}{4},
\end{equation}
where $\alpha(S)$ is the angle of holomorphy, defined as the smallest angle between two nonzero vectors from two linear subspaces of the underlying real vector space of $M$.

In particular, the complex hyperbolic space $\mathbb{H}_\mathbb{C}^n$ is a K\"{a}hler structure with $\kappa=-1$. Since $0\leq\cos^2\alpha(S)\leq1$, then from Eq. (\ref{eq:K(S)}), we can have $-1\leq K(S)\leq -1/4$ for any $2$-dimensional subspace $S\subset\mathcal{T}M$ of $\mathbb{H}_\mathbb{C}^n$, i.e., the (sectional) curvature is not constant in $\mathbb{H}_\mathbb{C}^n$, but pinched between $-1$ and $-1/4$. Thus we proved the non-constant curvature of $\mathbb{H}_\mathbb{C}^n$.

Specifically, we discuss the complex projective lines and totally real planes in the unit ball model of the complex hyperbolic space: $ \mathcal{B}_\mathbb{C}^n=\{(z_1,\cdots,z_n,1)||z_1|^2+\cdots+|z_n|^2<1\}$.

First let's consider the case of complex projective lines. Consider a complex line $L$ in $\mathbb{C}^n$ that intersects the unit ball model $\mathcal{B}_\mathbb{C}^n$. Let $\mathbf{z}$ be any point in $L\cap\mathcal{B}_\mathbb{C}^n$. We can apply an element of $\text{PU}(n,1)$ to $L$ so that it becomes the last coordinate axis $\{(\mathbf{0},z_n)|z_n\in\mathbb{C}\}$, whose intersection with $\mathcal{B}_\mathbb{C}^n$ is the disk $|z_n|<1$. Then the restriction of the Bergman metric to this disc is the Poincar\'e metric~\citep{beardon2012geometry} of constant curvature $-1$.

In order to see this, let $\mathbf{z}=(\mathbf{0},z_n,1)$ and $\mathbf{w}=(\mathbf{0},w_n,1)$, $\mathbf{z},\mathbf{w}\in L\cap\mathcal{B}_\mathbb{C}^n$, then from Eq. (9) in Section \ref{sec:unit ball model}, the distance between $\mathbf{z}$ and $\mathbf{w}$ is given by Eq. (\ref{eq:distance}). 
Then we have
\begin{equation}
    \cosh^2(\frac{d_{\mathcal{B}_\mathbb{C}^n}(\mathbf{z},\mathbf{w})}{2})=\frac{\llangle\mathbf{z},\mathbf{w}\rrangle\llangle\mathbf{w},\mathbf{z}\rrangle}{\llangle\mathbf{z},\mathbf{z}\rrangle\llangle\mathbf{w},\mathbf{w}\rrangle}=\frac{|z_n\overline{w_n}-1|^2}{(|z_n|^2-1)(|w_n|^2-1)},
\end{equation}
which is just the Poincar\'e metric~\citep{beardon2012geometry}.

Next consider a totally real plane $p$. Any totally real plane $p$ is the image under an element of $\text{PU}(n,1)$ of the subspace comprising those points of $\mathcal{B}_\mathbb{C}^n$ with real coordinates, that is actually an embedded copy of the real hyperbolic space $\mathbb{H}_\mathbb{R}^n=\{(x_1,\dots,x_n)|x_1,\dots,x_n\in\mathbb
R\}$. This subspace intersects $\mathcal{B}_\mathbb{C}^n$ in the subset consisting of those points with $x_1^2+\dots+x_n^2<1$. Then the Bergman metric restricted to this real-space unit ball is just the Klein-Beltrami metric~\citep{ratcliffe1994foundations} on the unit ball in $\mathbb{R}^n$ with constant curvature $-1/4$.

To see this, let $\mathbf{x}=(x_1,\dots,x_n,1)$ and $\mathbf{y}=(y_1,\dots,y_n,1)$, $\mathbf{x},\mathbf{y}\in\mathbb{H}_\mathbb{R}^n\cap\mathcal{B}_\mathbb{C}^n$, then apply the similar process with the above, we have
\begin{equation}
    \cosh^2(\frac{d_{\mathcal{B}_\mathbb{C}^n}(\mathbf{x},\mathbf{y})}{2})=\frac{\llangle\mathbf{x},\mathbf{y}\rrangle\llangle\mathbf{y},\mathbf{x}\rrangle}{\llangle\mathbf{x},\mathbf{x}\rrangle\llangle\mathbf{y},\mathbf{y}\rrangle}=\frac{(x_1y_1+\dots+x_ny_n-1)^2}{(x_1^2+\dots+x_n^2-1)(y_1^2+\dots+y_n^2-1)},
\end{equation}
which is the Klein-Beltrami metric~\citep{ratcliffe1994foundations} on the unit ball in $\mathbb{R}^n$ with constant curvature $-1/4$.

Therefore, we proved that the curvature of $\mathbb{H}_\mathbb{C}^n$ is $-1$ in the directions of complex projective lines while $-1/4$ in the directions of totally real planes.
\end{proof}
\begin{remark}
Curvature in the complex hyperbolic space is a very complicated topic in geometric group theory and differential geometry. The complex projective lines and the totally real planes are two kinds of special subspaces, whose curvatures are presented above. For the subspace that lives in between the two special cases,
we refer the interested readers to~\citep{fishernotes} for an interesting example in the complex hyperbolic space (its curvature differs with our work with a constant multiplier $4$). In Section 5 and Figure 5 of~\citep{fishernotes}, the author explored the curvature of a triangle in complex hyperbolic geometry with numerical computation.
\end{remark}

The non-constant curvature, which we expect to be favorable for embedding various hierarchical structures, is one of the main differences between $\mathbb{H}_\mathbb{C}^n$ and $\mathbb{H}_\mathbb{R}^n$.

\subsubsection{Exponential Volume Growth}
The complex hyperbolic space also has the tree-like exponential volume growth property. The volume of a ball with radius $\rho$ in $\mathbb{H}_\mathbb{C}^n$ is given by
\begin{equation}
\label{eq:complex ball volume}
    vol(B_{\mathbb{H}_\mathbb{C}^n}(\rho))=\frac{8^n\sigma_{2n-1}}{2n}\sinh^{2n}(\rho/2)\sim e^{n\rho},
\end{equation}
where $\sigma_{2n-1}=2\pi^n/n!$ is the Euclidean volume of the unit sphere $S^{2n-1}\in\mathbb{C}^n$. 

\begin{table}[t]
    \caption{Comparison of Euclidean, hyperbolic, and complex hyperbolic geometries. $n$ denotes the dimensionality and $\rho$ denotes the radius of the ball.}
\label{tab:geometries}
\begin{center}
\begin{small}
\begin{tabular}{l|l|l}
\toprule
Geometry                 & Curvature    & Volume growth of ball   \\
                 \midrule
Euclidean        & $0$ (constant)            & $\sim\rho^n$ (polynomially) \\
Hyperbolic & $K<0$ (constant negative) & $\sim e^{\sqrt{-K}(n-1)\rho}$ (exponentially)  \\
Complex hyperbolic & pinched between $-1$ and $-\frac{1}{4}$ (variable negative) & $\sim e^{n\rho}$ (exponentially)  \\
\bottomrule
\end{tabular}
\end{small}
\end{center}
\end{table}

From the properties of the complex hyperbolic geometry, we expect that the complex hyperbolic space can naturally handle data with diverse local structures in virtue of the variable curvature as presented in Theorem \ref{theo:complex hyperbolic curvature} while preserving the tree-like properties as shown in Eq. (\ref{eq:complex ball volume}). 
In summary, Table \ref{tab:geometries} outlines our concerned properties of Euclidean, hyperbolic, and complex hyperbolic geometry.

From this section, we see that complex hyperbolic geometry and hyperbolic geometry are typically of different characteristics. The $n$-dimensional ($n$-d) complex hyperbolic space is not simply the $2n$-d hyperbolic space or the product of two $n$-d hyperbolic spaces. This implies that our complex hyperbolic embedding model is intrinsically different from the hyperbolic embedding methods~\citep{DBLP:conf/nips/NickelK17,DBLP:conf/icml/NickelK18} or the product manifold embeddings~\citep{DBLP:conf/iclr/GuSGR19}.

\section{Unit Ball Embeddings}
\label{sec:approach}
We propose to embed the hierarchically structured data into the unit ball model of the complex hyperbolic space. In this section, we introduce our approach in detail.

\subsection{Embeddings in the Unit Ball Model}
\label{sec:loss function}
Given the hierarchical data containing a set of nodes $X=\{x_p\}_{p=1}^m$ and a set of edges $E=\{(x_p,x_q)|x_p,x_q\in X\}$, we aim to learn the embeddings of the nodes $\mathbf{Z}=\{\mathbf{z}_p\}_{p=1}^m$, where $\mathbf{z}_p\in\mathcal{B}_\mathbb{C}^n=\{(z_1,\cdots,z_n,1)||z_1|^2+\cdots+|z_n|^2<1\}$ (Eq. (\ref{eq:unit ball})). 

The objective of the embeddings is to recover the structures of input data, including the distances between the nodes as well as the partial order in the hierarchies. Here we adopt the soft ranking loss used in the Poincar\'e ball embeddings~\citep{DBLP:conf/nips/NickelK17} and the hyperboloid embeddings~\citep{DBLP:conf/icml/NickelK18}, which aims at preserving the hierarchical relationships among nodes:
\begin{equation}
\label{eq:soft ranking loss}
    L=\sum_{(x_p,x_q)\in E}\log\frac{e^{-d_{\mathcal{B}_\mathbb{C}^n}(\mathbf{z}_p,\mathbf{z}_q)}}{\sum_{x_k\in\mathcal{N}(x_p)}e^{-d_{\mathcal{B}_\mathbb{C}^n}(\mathbf{z}_p,\mathbf{z}_k)}},
\end{equation}
where $\mathcal{N}(x_p)=\{x_k:(x_p,x_k)\notin E\}\cup\{x_p\}$ is the set of negative examples for $x_p$ together with $x_p$. $d_{\mathcal{B}_\mathbb{C}^n}$ is the distance function in the unit ball model given in Eq. (\ref{eq:distance}).
The minimization of $L$ makes the connected nodes closer in the embedding space than those with no observed edges.

The learning process implicitly aligns the geometric structures of the embedding space and the underlying graph structures of data since the loss function aims at preserving the hierarchical relationships among nodes while the underlying graph structures are reflected by the hierarchical relationships. We learn the embeddings in the unit ball model, where the variable negative curvature of the complex hyperbolic space provides the capacity to deal with more varying structures. The experiments in Section \ref{sec:experiments} exhibit that the unit ball model learns the high-quality embeddings and captures the various hierarchical structures.

\subsection{Riemannian Optimization in the Unit Ball Model}
\label{sec:RSGD}
We learn the embeddings $\mathbf{Z}=\{\mathbf{z}_p\}_{p=1}^m$ through solving the optimization problem with constraint:
\begin{equation}
\label{eq:optimization problem}
    \mathbf{Z}\gets\arg\min_{\mathbf{Z}}L \qquad s.t. \forall\mathbf{z}_p\in\mathbf{Z}, \mathbf{z}_p\in\mathcal{B}_\mathbb{C}^n.
\end{equation}

For the optimization problems in Riemannian manifolds, \citet{DBLP:journals/tac/Bonnabel13} presented the Riemannian stochastic gradient descent (RSGD) algorithm, which we employ to optimize Eq. (\ref{eq:optimization problem}). To update an embedding $\mathbf{z}\in\mathcal{B}_\mathbb{C}^n$,\footnote{Here we omit the subscript of $\mathbf{z}_p$ for concision.} we need to obtain its Riemannian gradient $\nabla_R$. Specifically, 
the embedding is updated at the $t$-th iteration by $\mathbf{z}^{(t)}\gets\mathbf{z}^{(t-1)}-\eta^{(t)}\nabla_RL(\mathbf{z})$,
where $\eta^{(t)}$ is the learning rate at the $t$-th iteration and $\nabla_RL(\mathbf{z})$ is the Riemannian gradient of $L(\mathbf{z})$.

The Riemannian gradient $\nabla_R$ can be derived from rescaling the Euclidean gradient $\nabla_E$ with the inverse of the metric tensor $ds^2$ in Eq. (\ref{eq:metric tensor}). Apply the chain rule of differential functions and we have:
\begin{equation}
\label{eq:riemannian gradient}
    \nabla_RL(\mathbf{z})=\frac{1}{ds^2}\nabla_EL(\mathbf{z})=\frac{1}{ds^2}\frac{\partial L(\mathbf{z})}{\partial d_{\mathcal{B}_\mathbb{C}^n}(\mathbf{z},\mathbf{w})}\nabla_Ed_{\mathcal{B}_\mathbb{C}^n}(\mathbf{z},\mathbf{w}).
\end{equation}

$\frac{\partial L(\mathbf{z})}{\partial d_{\mathcal{B}_\mathbb{C}^n}(\mathbf{z},\mathbf{w})}$ is trivial to compute from Eq. (\ref{eq:soft ranking loss}).
In practical training, we implement and compute the complex hyperbolic embedding as its real part and imaginary part, i.e., $\mathbf{z}=\mathbf{x}+i\mathbf{y}$, where $i$ represents the \textit{imaginary unit}, i.e., $i^2=-1$. In order to get the gradient of the distance function $\nabla_Ed_{\mathcal{B}_\mathbb{C}^n}(\mathbf{z},\mathbf{w})$ in Eq. (\ref{eq:riemannian gradient}), we get the partial derivative with regard to the real part and the imaginary part, i.e., $\nabla_Ed_{\mathcal{B}_\mathbb{C}^n}(\mathbf{z},\mathbf{w})=\frac{\partial d_{\mathcal{B}_\mathbb{C}^n}(\mathbf{z},\mathbf{w})}{\partial\mathbf{x}}+i\frac{\partial d_{\mathcal{B}_\mathbb{C}^n}(\mathbf{z},\mathbf{w})}{\partial\mathbf{y}}$. The distance function in the unit ball model is given by Eq. (\ref{eq:distance}).
The full derivation of $\frac{\partial d_{\mathcal{B}_\mathbb{C}^n}(\mathbf{z},\mathbf{w})}{\partial\mathbf{x}}$ and $\frac{\partial d_{\mathcal{B}_\mathbb{C}^n}(\mathbf{z},\mathbf{w})}{\partial\mathbf{y}}$is as follows.

\begin{algorithm}[tb]
   \caption{RSGD of the unit ball embeddings.}
   \label{alg:RSGD}
\begin{algorithmic}
   \STATE {\bfseries Input:} initialization $\mathbf{z}^{(0)}$, number of iterations $T$, learning rates $\{\eta^{(t)}\}_{t=1}^T$.
   \FOR{$t=1$ {\bfseries to} $T$}
   \STATE Compute $\frac{\partial d_{\mathcal{B}_\mathbb{C}^n}}{\partial {\mathbf{x}}}$ and $\frac{\partial d_{\mathcal{B}_\mathbb{C}^n}}{\partial {\mathbf{y}}}$ by Eqs. (\ref{eq:real derivative}) and (\ref{eq:image derivative}).
   \STATE Compute $\nabla_EL(\mathbf{z})$ and $\nabla_RL(\mathbf{z})$ by Eq. (\ref{eq:riemannian gradient}).
   \STATE Update $\mathbf{z}^{(t)}$ by Eq. (\ref{eq:proj update}).
   \ENDFOR
\end{algorithmic}
\end{algorithm}

First, we need to introduce Wirtinger derivatives~\citep{wirtinger1927formalen}, which constructs a differential calculus for differential functions on complex domains.
\begin{definition}[Wirtinger derivatives]
The partial derivatives of a (complex) function $f(z)$ of a complex variable $z=x+iy\in\mathbb{C},x,y\in\mathbb{R}$, with respect to $z$ and $\bar{z}=x-iy$, respectively, are defined as:
\begin{equation*}
    \frac{\partial f(z,\bar{z})}{\partial z}=\frac{1}{2}(\frac{\partial}{\partial x}-i\frac{\partial}{\partial y})f(z,\bar{z}), \qquad
    \frac{\partial f(z,\bar{z})}{\partial\bar{z}}=\frac{1}{2}(\frac{\partial}{\partial x}+i\frac{\partial}{\partial y})f(z,\bar{z}).
\end{equation*}
\end{definition}

The Wirtinger derivatives can be rewritten as:
\begin{align}
    \label{eq:partial derivative with x}
    &\frac{\partial f(z,\bar{z})}{\partial x}=(\frac{\partial}{\partial z}+\frac{\partial}{\partial \bar{z}})f(z,\bar{z}), \\
    \label{eq:partial derivative with y}
    &\frac{\partial f(z,\bar{z})}{\partial y}=i(\frac{\partial}{\partial z}-\frac{\partial}{\partial \bar{z}})f(z,\bar{z}),
\end{align}

Let $p=\cosh(d_{\mathcal{B}_\mathbb{C}^n}(\mathbf{z},\mathbf{w}))=2\frac{\llangle\mathbf{z},\mathbf{w}\rrangle\llangle\mathbf{w},\mathbf{z}\rrangle}{\llangle\mathbf{z},\mathbf{z}\rrangle\llangle\mathbf{w},\mathbf{w}\rrangle}-1$, then $d_{\mathcal{B}_\mathbb{C}^n}(\mathbf{z},\mathbf{w})=arcosh(p)=\ln(p+\sqrt{p^2-1})$. Let $\mathbf{z}=(z_1,\dots,z_n,1)\in\mathcal{B}_\mathbb{C}^n$, then
\begin{align}
    \label{eq:partial derivative with z}
    \frac{\partial d_{\mathcal{B}_\mathbb{C}^n}(\mathbf{z},\mathbf{w})}{\partial z_j}=&\frac{\partial d_{\mathcal{B}_\mathbb{C}^n}(\mathbf{z},\mathbf{w})}{\partial p}\cdot\frac{\partial p}{\partial z_j}=\frac{1}{\sqrt{p^2-1}}\cdot\frac{\partial p}{\partial z_j}
    =\frac{2}{\sqrt{p^2-1}}\cdot\frac{\partial \frac{(z_1\overline{w_1}+\dots+z_n\overline{w_n}-1)\cdot\llangle\mathbf{w},\mathbf{z}\rrangle}{(z_1\overline{z_1}+\dots+z_n\overline{z_n}-1)\cdot\llangle\mathbf{w},\mathbf{w}\rrangle}}{\partial z_j} \nonumber \\
    =&\frac{2}{\sqrt{p^2-1}}\cdot\Big(\frac{\overline{w_j}\llangle\mathbf{w},\mathbf{z}\rrangle}{\llangle\mathbf{z},\mathbf{z}\rrangle\cdot\llangle\mathbf{w},\mathbf{w}\rrangle}-\frac{\overline{z_j}\llangle\mathbf{z},\mathbf{w}\rrangle\cdot\llangle\mathbf{w},\mathbf{z}\rrangle}{\llangle\mathbf{z},\mathbf{z}\rrangle^2\cdot\llangle\mathbf{w},\mathbf{w}\rrangle}\Big),
\end{align}
for $1\leq j\leq n$. Similarly, we can have
\begin{equation}
    \label{eq:partial derivative with conjugate z}
    \frac{\partial d_{\mathcal{B}_\mathbb{C}^n}(\mathbf{z},\mathbf{w})}{\partial \overline{z_j}}=\frac{2}{\sqrt{p^2-1}}\cdot\Big(\frac{{w_j}\llangle\mathbf{z},\mathbf{w}\rrangle}{\llangle\mathbf{z},\mathbf{z}\rrangle\cdot\llangle\mathbf{w},\mathbf{w}\rrangle}-\frac{{z_j}\llangle\mathbf{z},\mathbf{w}\rrangle\cdot\llangle\mathbf{w},\mathbf{z}\rrangle}{\llangle\mathbf{z},\mathbf{z}\rrangle^2\cdot\llangle\mathbf{w},\mathbf{w}\rrangle}\Big).
\end{equation}

Then by Eqs. (\ref{eq:partial derivative with x}), (\ref{eq:partial derivative with z}), and (\ref{eq:partial derivative with conjugate z}), we obtain
\begin{equation}
    \frac{\partial d_{\mathcal{B}_\mathbb{C}^n}}{\partial x_j}=\frac{\partial d_{\mathcal{B}_\mathbb{C}^n}(\mathbf{z},\mathbf{w})}{\partial z_j}+\frac{\partial d_{\mathcal{B}_\mathbb{C}^n}(\mathbf{z},\mathbf{w})}{\partial\overline{z_j}}=\frac{4}{\sqrt{p^2-1}}\Big(\frac{Re(\llangle\mathbf{z},\mathbf{w}\rrangle w_j)}{\llangle\mathbf{z},\mathbf{z}\rrangle\llangle\mathbf{w},\mathbf{w}\rrangle}-\frac{\llangle\mathbf{z},\mathbf{w}\rrangle\llangle\mathbf{w},\mathbf{z}\rrangle x_j}{\llangle\mathbf{z},\mathbf{z}\rrangle^2\llangle\mathbf{w},\mathbf{w}\rrangle}\Big).
\end{equation}

Similarly, by Eqs. (\ref{eq:partial derivative with y}), (\ref{eq:partial derivative with z}), and (\ref{eq:partial derivative with conjugate z}), we can get
\begin{equation}
    \frac{\partial d_{\mathcal{B}_\mathbb{C}^n}}{\partial y_j}=i(\frac{\partial d_{\mathcal{B}_\mathbb{C}^n}(\mathbf{z},\mathbf{w})}{\partial z_j}-\frac{\partial d_{\mathcal{B}_\mathbb{C}^n}(\mathbf{z},\mathbf{w})}{\partial\overline{z_j}})=\frac{4}{\sqrt{p^2-1}}\Big(\frac{Im(\llangle\mathbf{z},\mathbf{w}\rrangle w_j)}{\llangle\mathbf{z},\mathbf{z}\rrangle\llangle\mathbf{w},\mathbf{w}\rrangle}-\frac{\llangle\mathbf{z},\mathbf{w}\rrangle\llangle\mathbf{w},\mathbf{z}\rrangle y_j}{\llangle\mathbf{z},\mathbf{z}\rrangle^2\llangle\mathbf{w},\mathbf{w}\rrangle}\Big),
\end{equation}
where $Re(\cdot)$ and $Im(\cdot)$ denote the real and the imaginary part respectively. Then the partial derivatives of the unit ball model distance take the following formulas:
\begin{align}
\label{eq:real derivative}
    &\frac{\partial d_{\mathcal{B}_\mathbb{C}^n}}{\partial {\mathbf{x}}}=\frac{4}{\sqrt{p^2-1}}\Big(\frac{Re(\llangle\mathbf{z},\mathbf{w}\rrangle\mathbf{w})}{\llangle\mathbf{z},\mathbf{z}\rrangle\llangle\mathbf{w},\mathbf{w}\rrangle}-\frac{\llangle\mathbf{z},\mathbf{w}\rrangle\llangle\mathbf{w},\mathbf{z}\rrangle \mathbf{x}}{\llangle\mathbf{z},\mathbf{z}\rrangle^2\llangle\mathbf{w},\mathbf{w}\rrangle}\Big), \\
    \label{eq:image derivative}
    &\frac{\partial d_{\mathcal{B}_\mathbb{C}^n}}{\partial {\mathbf{y}}}=\frac{4}{\sqrt{p^2-1}}\Big(\frac{Im(\llangle\mathbf{z},\mathbf{w}\rrangle\mathbf{w})}{\llangle\mathbf{z},\mathbf{z}\rrangle\llangle\mathbf{w},\mathbf{w}\rrangle}-\frac{\llangle\mathbf{z},\mathbf{w}\rrangle\llangle\mathbf{w},\mathbf{z}\rrangle \mathbf{y}}{\llangle\mathbf{z},\mathbf{z}\rrangle^2\llangle\mathbf{w},\mathbf{w}\rrangle}\Big),
\end{align}

Since the embedding $\mathbf{z}$ should be constrained within the unit ball model, we apply the same projection strategy as~\citep{DBLP:conf/nips/NickelK17} via a small constant $\varepsilon$:
\begin{equation}
    proj(\mathbf{z})=
      \mathbf{z}/(|\mathbf{z}|-\varepsilon), \text{   if $|\mathbf{z}|\geq1$, else   } \mathbf{z}.
\end{equation}

To sum up, the update of $\mathbf{z}$ at the $t$-th iteration is
\begin{equation}
\label{eq:proj update}
    \mathbf{z}^{(t)}\gets proj\big(\mathbf{z}^{(t-1)}-\eta^{(t)}\nabla_RL(\mathbf{z})\big)= proj\big(\mathbf{z}^{(t-1)}-\eta^{(t)}\frac{1}{ds^2}\nabla_EL(\mathbf{z})\big).
\end{equation}

The RSGD steps of the unit ball embeddings are presented in Algorithm \ref{alg:RSGD}.

\section{Experiments}
\label{sec:experiments}

In experiments, we evaluate the performances of our approach and baselines on various hierarchical structures, including synthetic graphs and real-world data. We focus on the graph reconstruction task and the link prediction task.

\subsection{Experimental Settings}
\label{sec:experimental setting}
\subsubsection{Data}
\label{sec:data}
We use synthetic and real-world data that exhibit underlying hierarchical structures to evaluate our approach. The details are given in the following.

\begin{figure}[t]
\begin{center}
\centerline{\includegraphics[width=\columnwidth]{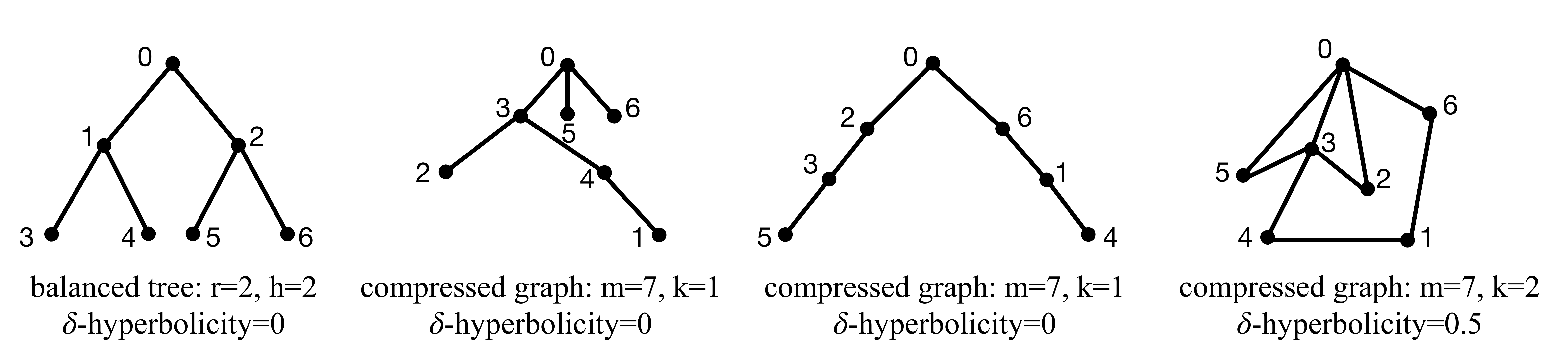}}
\caption{Simple examples of the synthetic data. The numbers $\{0,1,\dots,6\}$ represent the nodes. The compressed graph-($m=7$, $k=2$) on the right are aggregated from the middle two compressed graphs-($m=7$, $k=1$).}
\label{fig:synthetic}
\Description{There are four subfigures. The first subfigure shows a balanced tree with degree being 2 and depth being 2, while the hyperbolicity is 0. The second and the third subfigures are both compressed trees having 7 nodes, so their hyperbolicities are 0. The last subfigure is a compressed graph aggregated from the second and the third trees, and its hyperbolicity is 0.5.}
\end{center}
\end{figure}

\textbf{Synthetic.} We generate various balanced trees and compressed graphs using NetworkX package~\citep{SciPyProceedings_11}.\footnote{\url{https://networkx.org/documentation/stable/reference/generators.html}.} For \textbf{balanced trees}, we generate the balanced tree with degree $r$ and depth $h$. For \textbf{compressed graphs}, we generate $k$ random trees on $m$ nodes and then aggregate their edges to form a graph. We give some examples of the synthetic data in Figure \ref{fig:synthetic}. As we can see, the compressed graphs-($m=7$, $k=1$) are random trees on $7$ nodes, so their $\delta$-hyperbolicities are $0$. The compressed graph-($m=7$, $k=2$) is no longer a tree after aggregating from two trees. Its local structures are more varying and complicated.

\textbf{ICD10.} The $10$-th revision of International Statistical Classification of Diseases and Related Health Problems (ICD10)~\citep{bramer1988international} is a medical classification list provided by the World Health Organization.\footnote{\url{https://www.who.int/standards/classifications/classification-of-diseases}.}
We construct its full transitive closure as the ICD10 dataset.

\textbf{YAGO3-wikiObjects.} YAGO3~\citep{DBLP:conf/cidr/MahdisoltaniBS15} is a huge semantic knowledge base.\footnote{\url{https://yago-knowledge.org/}.}
It provides a taxonomy derived from Wikipedia and WordNet. We extract the Wikipedia concepts and entities that are descendants of $\langle wikicat\underline{~~}Objects\rangle$ as well as the hypernymy edges among them. We compute the transitive closure of the sampled taxonomy to construct the YAGO3-wikiObjects dataset.

\textbf{WordNet-noun.} WordNet~\citep{DBLP:journals/cacm/Miller95} is a large lexical database.\footnote{\url{https://wordnet.princeton.edu/}.} The hypernymy relation among all nouns forms a noun hierarchy. We use its full transitive closure as the WordNet-noun dataset.

\begin{figure}[t]
\begin{center}
{\includegraphics[width=\columnwidth]{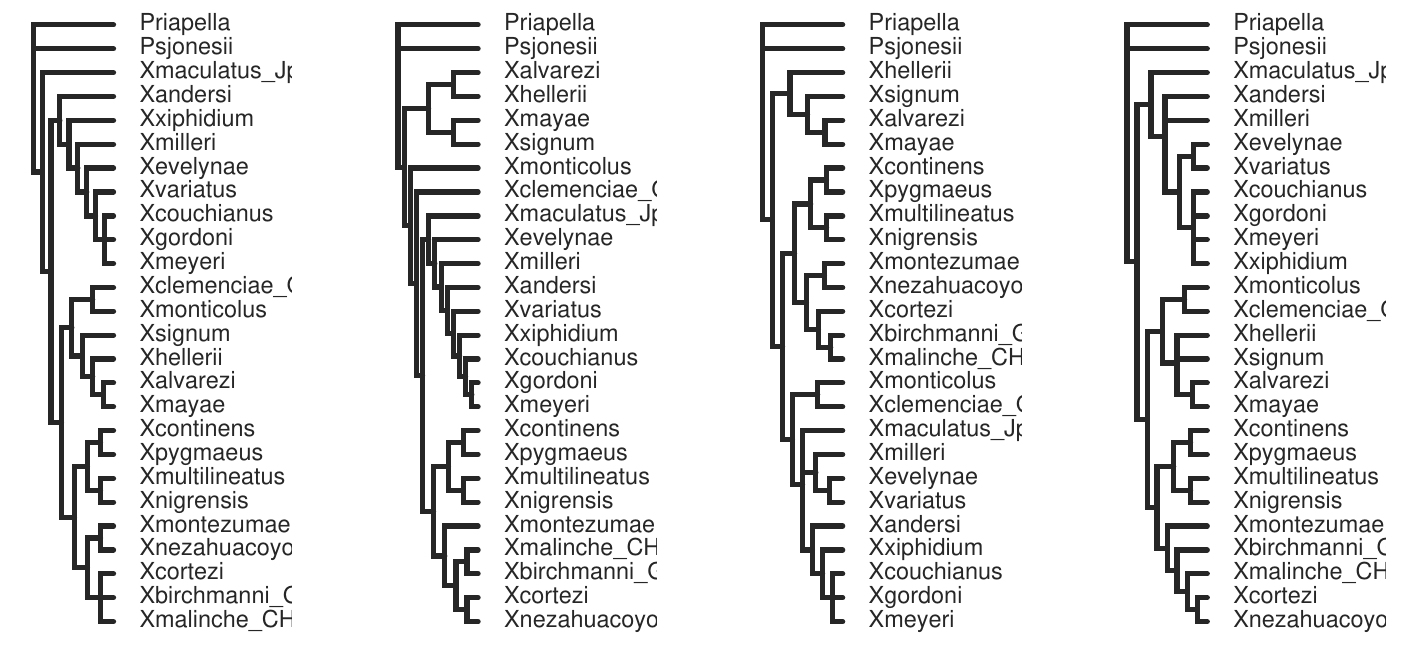}}
\caption{Some subtrees of Xiphophorus. The multitree dataset Xiphophorus contains $160$ trees on $26$ Xiphophorus fishes as leaf nodes.}
\label{fig:multitree tree list}
\Description{Fully described in the text.}
\end{center}
\end{figure}

\textbf{Xiphophorus.} The Xiphophorus is a multitree dataset from~\citep{cui2013data}, which contains $160$ trees representing mrbayes consensus trees inferred for different genomic regions on $26$ Xiphophorus fishes. Some examples of its subtrees are in Figure \ref{fig:multitree tree list} and its cloud tree plot is in Figure \ref{fig:more structures}. We use the saved MultiTree object in toytree package.\footnote{\url{https://toytree.readthedocs.io/en/latest/7-multitrees.html}.}

\begin{table}[t]
\caption{The real-world datasets statistics.}
    \label{tab:statistics}
    \begin{center}
    \begin{small}
    \begin{tabular}{l|r|r|r|r}
    \toprule
                     & Xiphophorus       & ICD10      & YAGO3-wikiObjects & WordNet-noun  \\
                            \midrule
Nodes        & 3,562         & 19,155     & 17,375           & 82,115      \\
Edges         & 7,536          & 78,357     & 153,643          & 743,086   \\
Depth        &  13          & 6          & 16               & 20      \\
Training edges   & 7,536     & 70,521     & 138,277          & 668,776     \\
Valid/Test edges  & 4,160     & 3,918      & 7,683            & 37,155     \\
$\delta$-hyperbolicity & 2.5 &      0.0            & 1.0 & 0.5 \\   
\bottomrule
    \end{tabular}
    \end{small}
    \end{center}
\end{table}

For the multitree (Xiphophorus), we use the full dataset as the training set and the edges containing the leaf nodes as the test set. For each real-world taxonomy (ICD10, YAGO3-wikiObjects, WordNet-noun), we randomly split the edges into train-validation-test sets with the ratio $90\%$:$5\%$:$5\%$. We make sure that any node in the validation and test sets must occur in the training set since otherwise, it cannot be predicted. But the edges in the validation and test sets do not occur in the training set since they are disjoint. We provide the statistics of the real-world datasets in Table \ref{tab:statistics}. The Gromov's $\delta$-hyperbolicity~\citep{gromov1987hyperbolic} measures the tree-likeness of graphs (refer to Section \ref{sec:delta-hyperbolicity} for definition). The lower $\delta$ corresponds to the more tree-like graph and trees have $0$ $\delta$-hyperbolicity.

\subsubsection{Tasks}
We evaluate the following two tasks:

\textbf{Graph reconstruction}: we train the embeddings of the full data and then reconstruct it from the embeddings. The task evaluates representation capacity.

\textbf{Link prediction}: we train the embeddings on the training set and predict the edges in the test set. The task evaluates generalization performance. 

\subsubsection{Baselines}
We compare our approach \textbf{UnitBall} to the following methods:
the sate-of-the-art combinatorial construction-based hyperbolic embedding method \textbf{TreeRep}~\citep{DBLP:conf/nips/SonthaliaG20},\footnote{\url{https://github.com/rsonthal/TreeRep}.} the \textbf{Product} hyperbolic embeddings~\citep{DBLP:conf/iclr/GuSGR19},\footnote{\url{https://github.com/HazyResearch/hyperbolics}.}
the optimization-based hyperbolic embeddings in the \textbf{Poincar\'e} ball model~\citep{DBLP:conf/nips/NickelK17} and the \textbf{Hyperboloid} model~\citep{DBLP:conf/icml/NickelK18},
the simple \textbf{Euclidean} embedding model.\footnote{\url{https://github.com/facebookresearch/poincare-embeddings}. The repository provides the implementation for Euclidean, Poincar\'e, and Hyperboloid.}
Note that Euclidean, Poincar\'e, Hyperboloid, and our approach UnitBall use the same loss function but learn in the different geometrical spaces. Therefore, the comparisons reveal the capacities of different geometrical models in different spaces.

For the baselines, we use their released codes to train the embeddings. 
For all methods, we tune the hyperparameters by grid search. For the graph reconstruction task, we tune the hyperparameters on balanced tree-(15,3) in $20$-dimensional embedding spaces ($10$-dimensional complex hyperbolic space for UnitBall), while for the link prediction task, we tune the hyperparameters on the validation sets in $32$-dimensional embedding spaces ($16$-dimensional complex hyperbolic space for UnitBall). In all experiments, we report the mean results over $5$ running executions.

The $n$-d complex hyperbolic embeddings have around double parameters of the $n$-d real embeddings since the $n$-d complex hyperbolic vectors have $n$-d real part and $n$-d imaginary part. For a fair comparison, in each experimental setting, we compare our $n$-d complex hyperbolic embeddings of UnitBall against the $2n$-d embeddings of the baselines. The results will also demonstrate that the $n$-d complex hyperbolic space is not simply the $2n$-d hyperbolic space, they have different capacities.

\subsubsection{Evaluation}
\label{sec:evaluation}
Our evaluation closely follows the setting of~\citep{DBLP:conf/nips/NickelK17,DBLP:conf/icml/NickelK18}, which infers the hierarchies from distances in the embedding space. We use the mean average precision (\textbf{MAP}), mean reciprocal rank (\textbf{MRR}), and \textbf{Hits@N} as our {evaluation metrics}, which are widely used for evaluating ranking and link prediction. Specifically, for each test edge $(z,w)$, we compute the distance between the embeddings $d_{\mathcal{B}_\mathbb{C}^n}(\mathbf{z},\mathbf{w})$ and rank it among the distances of all unobserved edges for $z$: $\{d_{\mathcal{B}_\mathbb{C}^n}(\mathbf{z},\mathbf{w}'):({z},{w}')\notin\text{Training}\}$. We then report the following evaluation metrics of the rankings. Denote $E_{test}$ as the test edge set and $V=\{z|\exists w, (z,w)\in E_{test}\}$ as the test node set. Let $NE_z=\{w_1,w_2,\dots,w_{|NE_z|}\}$ be the ground truth neighbor set of node $z$.

\paragraph{Mean average precision (MAP)} The average precision (AP) is a way to summarize the precision-recall curve into a single value representing the average of all precisions and the MAP score is calculated by taking the mean AP over all classes. For a node $z$, from the learned embeddings, we can obtain the nodes closest to its embedding $\mathbf{z}$. Let $R_{z,w_i}$ be the smallest set of such nodes that contains $w_i$ (the $i$-th neighbor of $z$). Then the MAP is defined as:
\begin{equation*}
    \text{MAP}=\frac{1}{|V|}\sum_{z\in V}\frac{1}{|NE_z|}\sum_{w_i\in NE_z}Precision(R_{z,w_i}).
\end{equation*}

\paragraph{Mean reciprocal rank (MRR)} The MRR is a statistic measure for evaluating a list of possible responses to a sample of queries, ordered by the probability of correctness. For a node $z$, from the learned embeddings, we can rank its distances with other nodes from the smallest to the largest. Let $rank_{w_i}$ be the rank of $w_i$ (the $i$-th neighbor of $z$). Then the MRR is defined as:
\begin{equation*}
    \text{MRR}=\frac{1}{|V|}\sum_{z\in V}\frac{1}{|NE_z|}\sum_{w_i\in NE_z}\frac{1}{rank_{w_i}}.
\end{equation*}

\paragraph{The proportion of correct types that rank no larger than $N$ (Hits@$N$)} Hits@$N$ measures whether the top $N$ predictions contain the ground truth labels. For a node $z$, from the learned embeddings, we can obtain the set of $N$ nodes closest to its embedding $\mathbf{z}$, denoted as $R_z^N$. Then the Hits@$N$ is defined as:
\begin{equation*}
    \text{Hits@}N=\frac{1}{|V|}\sum_{z\in V}\mathbb{I}(|R_z^N\cap NE_z|\geq 1),
\end{equation*}
where $\mathbb{I}(|R_z^N\cap NE_z|\geq 1)$ is the indicator function.

\subsubsection{Hardware}
\label{sec:hardware}
We conduct all the experiments except TreeRep on four NVIDIA GTX 1080Ti GPUs with $8$GB memory each. For TreeRep, we need more memory to store the distance matrices, so we use a $96$-core NVIDIA T4 GPU server with $503$GB memory.

\subsection{Experiments on Synthetic Data}
\label{sec:synthetic results}

\subsubsection{Graph Reconstruction Results on Balanced Trees}
\label{sec:balanced trees}
To compare the representation capacities of UnitBall and the hyperbolic embedding models for the tree structures, we evaluate the graph reconstruction task on the synthetic balanced trees. A balanced tree-$(r,h)$ has degree $r$ and depth $h$, so it has $r^0+\dots+r^d$ nodes and $r^0+\dots+r^d-1$ edges. The $\delta$-hyperbolicity of any balanced tree is $0$. We embed the balanced trees into $20$-d hyperbolic space for the baselines and $10$-d complex hyperbolic space for UnitBall. 

\begin{figure}[t]
\begin{center}
\centerline{\includegraphics[width=\columnwidth]{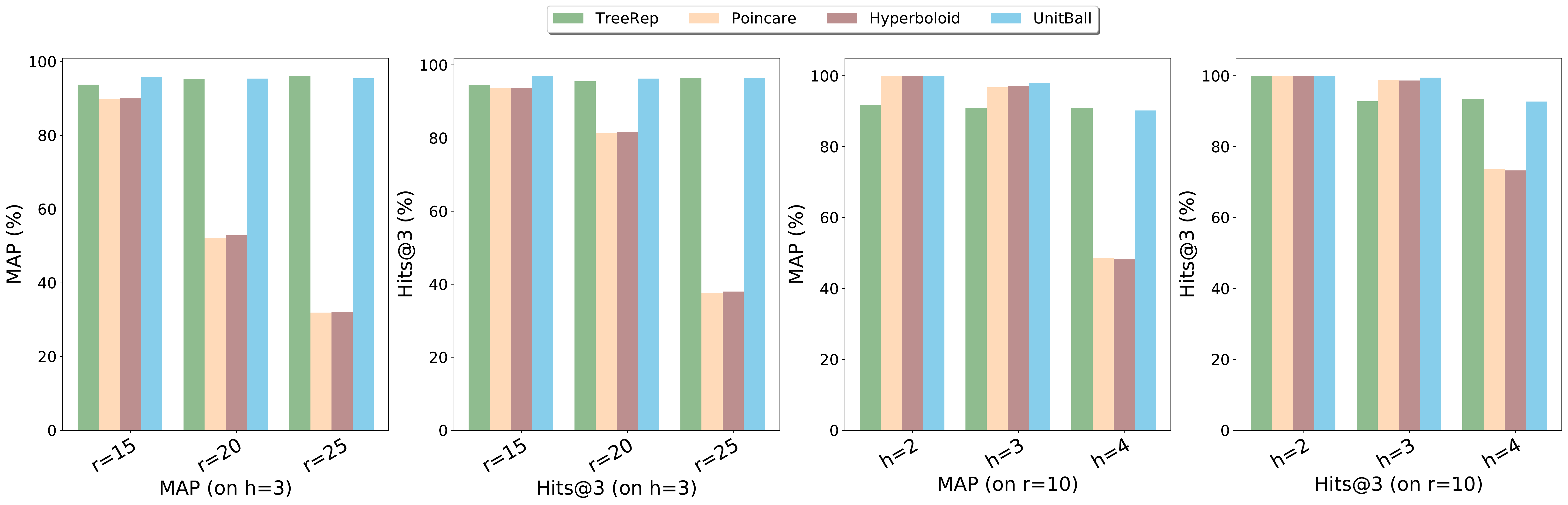}}
\caption{Evaluation of graph reconstruction on synthetic balanced trees in $20$-d embedding spaces ($10$-d complex hyperbolic space for UnitBall). $r$ represents the degree while $h$ represents the depth.}
\label{fig:tree}
\Description{Fully described in the text.}
\end{center}
\end{figure}

Figure \ref{fig:tree} presents the MAP and Hits@3 scores with varying $r$ and $h$. 
We see that when the tree is in small scale, e.g., $(r,h)=(15,3),(10,2),(10,3)$, all methods have very good performances, demonstrating the expected powerful capacities of hyperbolic geometry and complex hyperbolic geometry on tree structures. However, when the breadth or the depth increases, the performances of Poincar\'e and Hyperboloid drop rapidly, suggesting that the optimization-based embeddings in $\mathbb{H}_\mathbb{R}^{20}$ are not effective enough for reconstructing trees of such scales. 

In comparison, UnitBall and TreeRep achieve stable performances for larger trees.
TreeRep learns a tree structure from the data as an intermediate step and then embeds the learned trees into the hyperbolic space using Sarkar's construction~\citep{DBLP:conf/gd/Sarkar11}. When the input data is a tree, TreeRep exactly recovers the original tree structure. Figure \ref{fig:tree} shows that UnitBall achieves comparable or even better performances than TreeRep on the balanced trees. The results demonstrate that UnitBall does not compromise on trees. It produces high-quality embeddings for tree structures.


\begin{figure}[t]
\begin{center}
\begin{subfigure}
{\includegraphics[width=\columnwidth]{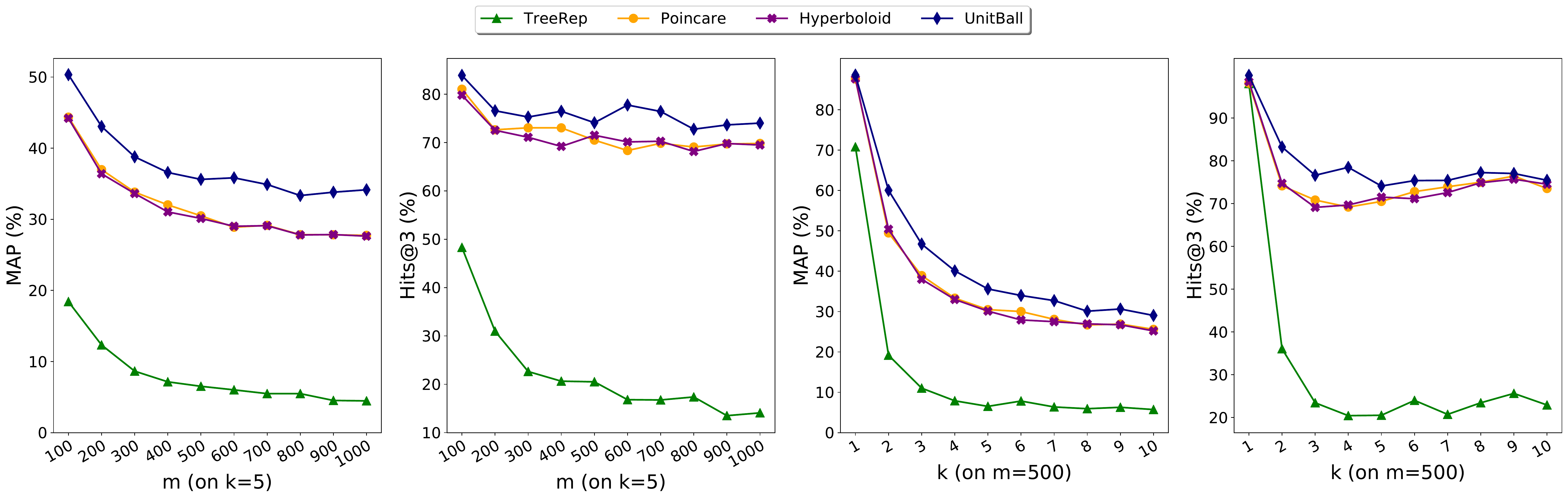}}
\end{subfigure}
\begin{subfigure}
{\includegraphics[width=0.45\columnwidth]{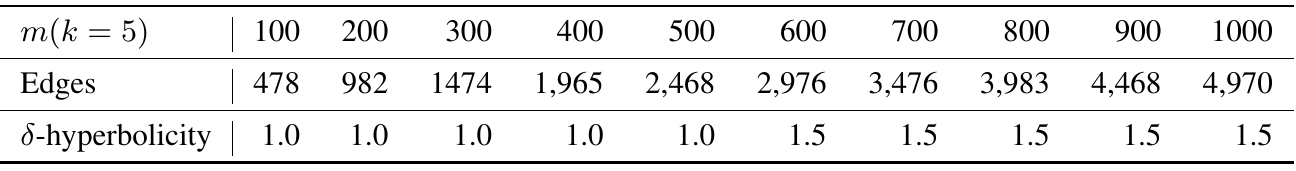}}
\end{subfigure}
\begin{subfigure}
{\includegraphics[width=0.45\columnwidth]{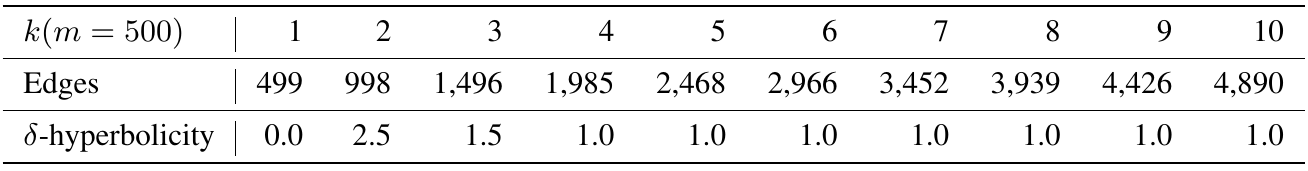}}
\end{subfigure}
\caption{Evaluation of graph reconstruction on synthetic compressed graphs in $20$-d embedding spaces ($10$-d complex hyperbolic space for UnitBall). $m$ represents the number of nodes in the graph while $k$ represents the number of random trees aggregated to the graph ($k$ controls the denseness and noise level of the graph). The statistics of the compressed graphs are provided in the tables.}
\label{fig:compressed_tree}
\Description{Fully described in the text.}
\end{center}
\end{figure}

\begin{table}[t]
\caption{Results of MAP and Hits@3 on graph reconstruction in synthetic compressed graphs. The best results are shown in boldface. The second best results are underlined.}
\label{tab:product embedding results}
    \begin{center}
    \begin{small}
    \begin{tabular}{l|rrrrrrrrrr}
    \toprule
     & \multicolumn{10}{c}{Compressed graph-$(m,k)$ (MAP)} \\
     \midrule
    k (m=500)                                            & 1     & 2     & 3     & 4     & 5     & 6     & 7     & 8     & 9     & 10    \\
    $\delta$-hyperbolicity & 0.0 & 2.5 & 1.5 & 1.0 & 1.0 & 1.0 & 1.0 & 1.0 & 1.0 & 1.0 \\
    \midrule
Product-$(\mathbb{H}_\mathbb{R}^2)^{16}$ & \underline{66.06} & \underline{7.60}  & \underline{7.55}  & 11.09 & \underline{20.81} & 24.56 & 24.07 & 21.62 & 18.79 & 16.16 \\
Product-$(\mathbb{H}_\mathbb{R}^4)^8$  & 65.77 & 7.14  & 7.24  & \underline{11.79} & 20.60 & \underline{24.94} & 22.85 & \underline{22.32} & 18.50 & 16.27 \\
Product-$(\mathbb{H}_\mathbb{R}^8)^4$  & 65.42 & 6.28  & 6.81  & 11.43 & 19.03 & 23.88 & \underline{24.99} & 20.50 & \underline{18.99} & \underline{16.45} \\
\midrule
UnitBall-$\mathbb{H}_\mathbb{C}^{16}$       & \textbf{84.72} & \textbf{52.74} & \textbf{44.73} & \textbf{39.75} & \textbf{35.32} & \textbf{33.17} & \textbf{32.48} & \textbf{29.13} & \textbf{29.86} & \textbf{28.58} \\
\bottomrule
  & \multicolumn{10}{c}{Compressed graph-$(m,k)$ (Hits@3)} \\
     \midrule
     Product-$(\mathbb{H}_\mathbb{R}^2)^{16}$ & \underline{80.34} & 8.14  & 9.84  & 14.81 & \underline{37.45} & 45.97 & 47.89 & \underline{47.19} & \underline{40.85} & 38.88 \\
Product-$(\mathbb{H}_\mathbb{R}^4)^8$  & \underline{80.34} & \underline{8.35}  & 9.43  & 16.23 & 36.23 & \underline{47.98} & 44.87 & \underline{47.19} & 39.84 & 36.87 \\
Product-$(\mathbb{H}_\mathbb{R}^8)^4$  & 79.36 & 6.21  & \underline{10.25} & \underline{16.63} & 31.38 & 44.15 & \underline{49.50} & 43.57 & 40.64 & \underline{40.08} \\
\midrule
UnitBall-$\mathbb{H}_\mathbb{C}^{16}$     & \textbf{97.71} & \textbf{75.87} & \textbf{72.61} & \textbf{74.92} & \textbf{73.21} & \textbf{73.12} & \textbf{76.12} & \textbf{76.57} & \textbf{75.72} & \textbf{74.08} \\
\bottomrule
     & \multicolumn{10}{c}{Compressed graph-$(m,k)$ (MAP)} \\
     \midrule
m (k=5)                                   & 100   & 200   & 300   & 400   & 500   & 600   & 700   & 800   & 900   & 1000  \\
$\delta$-hyperbolicity & 1.0 & 1.0 & 1.0 & 1.0 & 1.0 & 1.5 & 1.5 & 1.5 & 1.5 & 1.5 \\
    \midrule
Product-$(\mathbb{H}_\mathbb{R}^2)^{16}$ & 30.53 & \underline{29.15} & \underline{25.72} & 21.74 & \underline{20.81} & \underline{19.34} & 19.07 & 16.63 & \underline{15.59} & \underline{14.82} \\
Product-$(\mathbb{H}_\mathbb{R}^4)^8$    & \underline{32.03} & 27.63 & 23.82 & \underline{22.32} & 20.60 & 17.94 & \underline{19.12} & \underline{17.36} & 15.39 & 14.74 \\
Product-$(\mathbb{H}_\mathbb{R}^8)^4$    & 31.39 & 27.95 & 23.17 & 19.90 & 19.03 & 17.96 & 18.09 & 16.13 & 13.86 & 13.47 \\
UnitBall-$\mathbb{H}_\mathbb{C}^{16}$    & \textbf{47.80} & \textbf{40.93} & \textbf{38.52} & \textbf{35.81} & \textbf{35.32} & \textbf{35.68} & \textbf{34.73} & \textbf{34.69} & \textbf{34.92} & \textbf{34.88} \\
\bottomrule
 & \multicolumn{10}{c}{Compressed graph-$(m,k)$ (Hits@3)} \\
     \midrule
Product-$(\mathbb{H}_\mathbb{R}^2)^{16}$ & 38.78 & \underline{52.02} & \underline{43.58} & 35.61 & \underline{37.45} & \underline{34.29} & 34.54 & 28.34 & 27.44 & 24.77 \\
Product-$(\mathbb{H}_\mathbb{R}^4)^8$    & 46.94 & 43.43 & 39.53 & \underline{38.38} & 36.23 & 29.58 & \underline{35.84} & 29.35 & \underline{27.55} & \underline{27.60} \\
Product-$(\mathbb{H}_\mathbb{R}^8)^4$    & \underline{54.08} & 42.42 & 35.47 & 33.59 & 31.38 & 32.44 & 30.49 & \underline{29.72} & 24.64 & 24.57 \\
UnitBall-$\mathbb{H}_\mathbb{C}^{16}$    & \textbf{84.35} & \textbf{75.42} & \textbf{77.36} & \textbf{72.14} & \textbf{73.21} & \textbf{73.00} & \textbf{73.99} & \textbf{72.63} & \textbf{72.23} & \textbf{74.01} \\
\bottomrule
    \end{tabular}
    \end{small}
    \end{center}
\end{table}

\subsubsection{Graph Reconstruction Results on Compressed Graphs}
\label{sec:results on compressed graphs}
To illustrate the benefits of UnitBall on varying hierarchical structures, we evaluate on the synthetic compressed graphs. The compressed graphs have local tree structures while being much more complicated than trees. Each compressed graph-$(m,k)$ consists of $m$ nodes and is aggregated from $k$ random trees on the $m$ nodes. The bigger $k$ corresponds to the denser and noisier graph. 

Figure \ref{fig:compressed_tree} depicts that the graph reconstruction results drop down with the increase of $m$ and $k$, which represents the increase of graph scale and denseness respectively. Remarkably, UnitBall outperforms all baselines on the challenging data, showing that UnitBall handles the noisy locally tree-like structures better. 
TreeRep has comparable results with other methods when $(m,k)=(500,1)$ since when $k=1$, the graph is exactly a tree, i.e.,  $\delta=0$. However, when $k>1$ and $\delta>0$, the data metrics deviate from tree metrics, in which case it does not help much to learn a tree structure from the data as an intermediate step.

As mentioned in Section \ref{sec:related work}, the product space embeddings~\citep{DBLP:conf/iclr/GuSGR19} tackles the challenges in varying local structures by jointly learning the curvature and the embeddings of data in a product manifold. Although it is impractical to search for the best manifold combination among enormous combinations for each new structure, it is worth exploring the comparisons between the complex hyperbolic embeddings and products of hyperbolic embeddings. Therefore, we also evaluate the $16$-dimensional UnitBall complex hyperbolic embeddings and $32$-dimensional product hyperbolic embeddings on $(\mathbb{H}_\mathbb{R}^2)^{16}$, $(\mathbb{H}_\mathbb{R}^4)^8$, $(\mathbb{H}_\mathbb{R}^8)^4$. The results are reported in Table \ref{tab:product embedding results}.

When $k=1$ ($\delta=0$), UnitBall and the product hyperbolic embeddings both have much better performances. When $k>1$, UnitBall still outperforms the product hyperbolic embeddings by a large margin. Especially when $m=500, k=2, 3$, the $\delta$ is big, which means the graph deviates from tree structures a lot, the product hyperbolic embeddings fail to reconstruct the graph while UnitBall successfully handles the noisy structures. 

\subsection{Experiments on Real-World Data}
\label{sec:real-world results}

\begin{table}[t]
\caption{Evaluation of taxonomy link prediction in $32$-d embedding spaces ($16$-d complex hyperbolic space for UnitBall). The best results are shown in boldface. The second best results are underlined.}
    \label{tab:overall results}
    \begin{center}
    \begin{small}
    \begin{tabular}{l|rrr|rrr|rrr}
    \toprule
                     & \multicolumn{3}{c|}{ICD10}                              & \multicolumn{3}{c|}{YAGO3-wikiObjects}                        & \multicolumn{3}{c}{WordNet-noun}                      \\
            & MAP         & MRR          & Hits@3      & MAP         & MRR      & Hits@3      & MAP         & MRR    & Hits@3      \\
            \midrule
{Euclidean}   & 3.75  & 3.72  & 2.39  & 4.85  & 4.45  &  2.78  & 5.59  & 5.36   & 3.16  \\
TreeRep & 4.96 & 7.92 & 8.49 & 20.19 & 21.85 & 27.19 & 9.30 & 9.98 & 11.90 \\
{Poincar\'e}    & \underline{35.24} & \underline{34.45} & 52.71 & 30.06 & 28.47 & 41.61 & 25.46 & 23.99 &  \underline{27.80} \\
{Hyperboloid} & 34.80 & 34.01 & \underline{52.88} & \underline{30.80} & \underline{29.21} &  \underline{43.17} & \underline{25.65} & \underline{24.15} &  27.50 \\
{UnitBall}   & \textbf{47.88} & \textbf{46.96} &  \textbf{70.28} & \textbf{33.33} & \textbf{31.85}  & \textbf{47.41} & \textbf{27.29} & \textbf{25.93}  & \textbf{32.95} \\
\bottomrule
    \end{tabular}
    \end{small}
    \end{center}
\end{table}

\begin{table}[t]
\caption{Evaluation of graph reconstruction on the real-world taxonomies (the dimension is $32$ for TreeRep and $16$ for UnitBall). For memory cost, the unit is \textit{GiB}.}
    \label{tab:taxonomy reconstruction results}
    \begin{center}
    \begin{small}
    \begin{tabular}{l|rrr|rrr|rrr}
    \toprule
         & \multicolumn{3}{c|}{ICD10} & \multicolumn{3}{c|}{YAGO3-wikiObjects} & \multicolumn{3}{c}{WordNet-noun} \\
         & MRR   & Hits@1 & Memory & MRR   & Hits@1 & Memory & MRR   & Hits@1 & Memory \\
         \midrule
TreeRep  & 26.74 & 91.97  & 30           & 36.71 & 95.39  & 21           & 16.99 & 90.51  & 226          \\
UnitBall & 47.47 & 98.93  & 0.005        & 39.65 & 96.10  & 0.005        & 28.88 & 94.95  & 0.02                 \\
\bottomrule
    \end{tabular}
    \end{small}
    \end{center}
\end{table}

\begin{table}[t]
\caption{Evaluation of taxonomy link prediction in different embedding dimensions (the embedding dimension for UnitBall is half of other models). The best results are shown in boldface. The second best results are underlined. TreeRep is not applicable to $128$-d WordNet-noun due to the large memory cost, so we do not include the results.}
\label{tab:dimension}
\begin{center}
\begin{small}
\begin{tabular}{l|rrr|rrr|rrr}
\toprule
 & \multicolumn{9}{c}{ICD10}                                        \\
            & \multicolumn{3}{c}{$8$-dimensional} & 
            \multicolumn{3}{c}{$32$-dimensional} &
            \multicolumn{3}{c}{$128$-dimensional} \\
            \midrule
            & MAP       & MRR       & Hits@3  & MAP       & MRR       & Hits@3  & MAP        & MRR       & Hits@3     \\
Euclidean   & 2.57 & 2.57 & 1.32 & 3.75 & 3.72 & 2.39 & 10.83 & 10.48 & 4.66      \\
TreeRep & 3.44 & 3.90 & 6.03 & 4.96 & 7.92 & 8.49 & 8.09 & 8.74 & 17.23 \\
Poincar\'e    & \underline{35.73} & \underline{34.94} & \underline{53.10} & \underline{35.24} & \underline{34.45} & 52.71 & 34.47 & 33.70 & 52.19      \\
Hyperboloid & 35.56 & 34.77 & 51.90 & 34.80 & 34.01 & \underline{52.88}  & \underline{34.93} & \underline{34.15} & \underline{52.98}    \\
UnitBall   & \textbf{44.05} & \textbf{43.26} & \textbf{61.54}  & \textbf{47.88} & \textbf{46.96} & \textbf{70.28} & \textbf{46.54} & \textbf{45.59} & \textbf{70.03}  \\
\bottomrule
                        & \multicolumn{9}{c}{YAGO3-wikiObjects}                                    \\
            &  \multicolumn{3}{c}{$8$-dimensional} & \multicolumn{3}{c}{$32$-dimensional} & \multicolumn{3}{c}{$128$-dimensional} \\
            \midrule
            & MAP       & MRR       & Hits@3    & MAP        & MRR       & Hits@3     & MAP       & MRR       & Hits@3    \\
Euclidean   & 1.02      & 0.92      & 0.57  & 4.85 & 4.45 & 2.78  & 16.67      & 15.76     & 15.97      \\
TreeRep & 16.91 & 17.48 & 27.53 & 20.19 & 21.85 & 27.19 & 21.18 & 23.44 & 32.84  \\
Poincar\'e    & 29.70     & 28.13     & 41.64  & 30.06 & 28.47 & 41.61   & 29.93      & 28.35     & 41.53      \\
Hyperboloid & \underline{30.87} & \underline{29.28} & \underline{43.50} & \underline{30.80} & \underline{29.21} & \underline{43.17} & \underline{30.68} & \underline{29.07} & \underline{42.86}     \\
UnitBall    & \textbf{31.40}     & \textbf{29.98}     & \textbf{44.25}   & \textbf{33.33} & \textbf{31.85} & \textbf{47.41} & \textbf{32.76}      & \textbf{31.28}     & \textbf{46.25}    \\
\bottomrule
            & \multicolumn{9}{c}{WordNet-noun}                                        \\
            & \multicolumn{3}{c}{$8$-dimensional} & 
            \multicolumn{3}{c}{$32$-dimensional} &
            \multicolumn{3}{c}{$128$-dimensional} \\
            \midrule
            & MAP       & MRR       & Hits@3  & MAP       & MRR       & Hits@3  & MAP        & MRR       & Hits@3     \\
Euclidean   & 1.07      & 1.05      & 0.63  & 5.59 & 5.36 & 3.16    & 14.33      & 13.35     & 8.82       \\
Poincar\'e    & \underline{25.23}     & \underline{23.78}     & 27.63 & 25.46 & 23.99 & \underline{27.80}    & 25.33      & 23.86     & 27.41      \\
Hyperboloid & \textbf{25.73}     & \textbf{24.24}     & \underline{27.67} & \underline{25.65} & \underline{24.15} & 27.50   & \underline{25.77}      & \underline{24.27}     & \underline{27.65}      \\
UnitBall   & 24.91     & 23.76     & \textbf{30.27} & \textbf{27.29} & \textbf{25.93} & \textbf{32.95}    & \textbf{27.29}      & \textbf{25.91}     & \textbf{32.77}      \\
\bottomrule
\end{tabular}
\end{small}
\end{center}
\end{table}

\subsubsection{Results on the Real-World Taxonomies}
\label{sec:overall results}
In this section, we evaluate the performances on the link prediction task for real-world taxonomies. Table \ref{tab:overall results} presents the results in $32$-d embedding spaces for baselines and $16$-d complex hyperbolic space for UnitBall.
Predicting missing links requires generalization capacity, and UnitBall still has the best performances on all three datasets.
Besides, we see that Euclidean shows shortages on these hierarchically-structured data, which is consistent with the results in previous works~\citep{DBLP:conf/nips/NickelK17,DBLP:conf/icml/NickelK18}. Similar to the results on the graph reconstruction task, Poincar\'e and Hyperboloid have very close performances, while Hyperboloid has slightly better results. They have significant improvements over Euclidean, but they still fall behind UnitBall, which demonstrates our claims that the non-constant negative curvature of the complex hyperbolic space addresses the varying hierarchical structures on real-world taxonomies. 

We notice that TreeRep does not perform well on the link prediction task. As mentioned in Section \ref{sec:related work}, the combinatorial construction-based embedding methods target minimizing the reconstruction distortion of data. However, minimizing the reconstruction distortion may overfit the training set, thus resulting in the unpromising generalization performance for unobserved edges. Hence, they are more suitable to learn the representation of graph data without missing links, such as the graph reconstruction tasks in Section \ref{sec:synthetic results}. Therefore, here we compare UnitBall with TreeRep on the real-world taxonomy reconstruction task. The results are presented in Table \ref{tab:taxonomy reconstruction results}. Its performance is much better than that on the link prediction task. In addition, UnitBall still outperforms TreeRep on reconstructing real-world taxonomies.

We also notice the memory issues of the combinatorial construction-based embedding methods. Although TreeRep is very efficient in embedding tree structures since it does not need the gradient-based optimization steps, it costs more memory resources for constructing the tree structures from data. It is basically a computation time vs. memory cost trade-off issue. For a graph with $m$ nodes, TreeRep needs to construct a matrix of size $c\cdot m\times c\cdot m$ to construct the tree structure, where $1\leq c\leq 2$ is a hyperparameter. We report the memory cost (\textit{GiB}) in Table \ref{tab:taxonomy reconstruction results}. UnitBall costs much less memory to learn the embeddings.

\subsubsection{Exploring the Embedding Dimensions}
\label{sec:dimension results}
In this section, we explore the link prediction performances in different embedding dimensions. The results are presented in Table \ref{tab:dimension}. We find that with the increase of the embedding dimension, Euclidean can have big improvements, but its performances in $128$-d still cannot surpass other methods in $8$-d. TreeRep also achieves better results with the increase of dimension, but overall its performances on the link prediction task are not very promising. By comparison, Poincar\'e, Hyperboloid, and UnitBall achieve great results steadily. 
$8$-d is already enough for Poincar\'e and Hyperboloid to handle the link prediction task while UnitBall has small improvements from $4$-d to $16$-d, then converges to the stable performance. 
Although on WordNet-noun, UnitBall in $4$-d has slightly lower MAP and MRR than Poincar\'e and Hyperboloid in $8$-d, it has much higher Hits@3.

\begin{figure}[t]
\begin{center}
\begin{subfigure}
{\includegraphics[width=0.45\columnwidth]{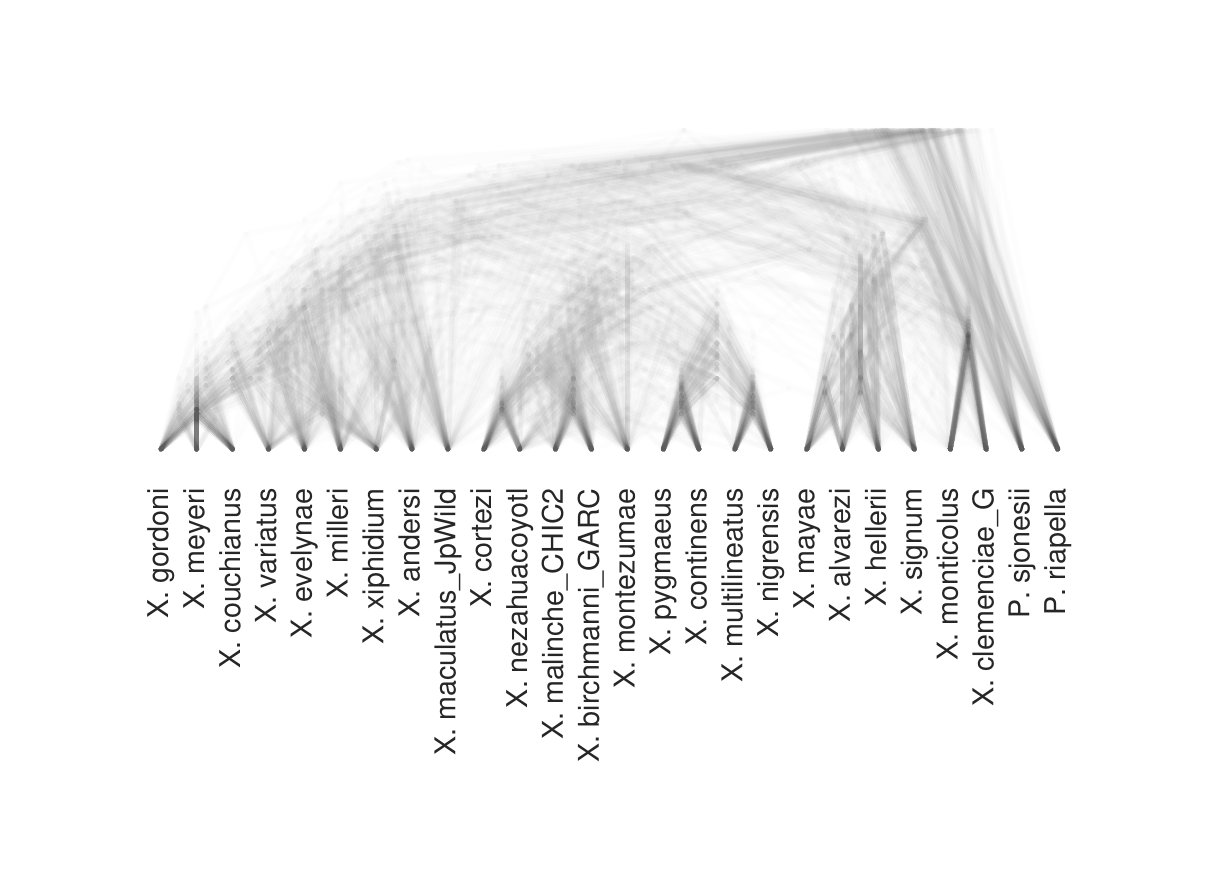}}
\end{subfigure}
\begin{subfigure}
{\includegraphics[width=0.45\columnwidth]{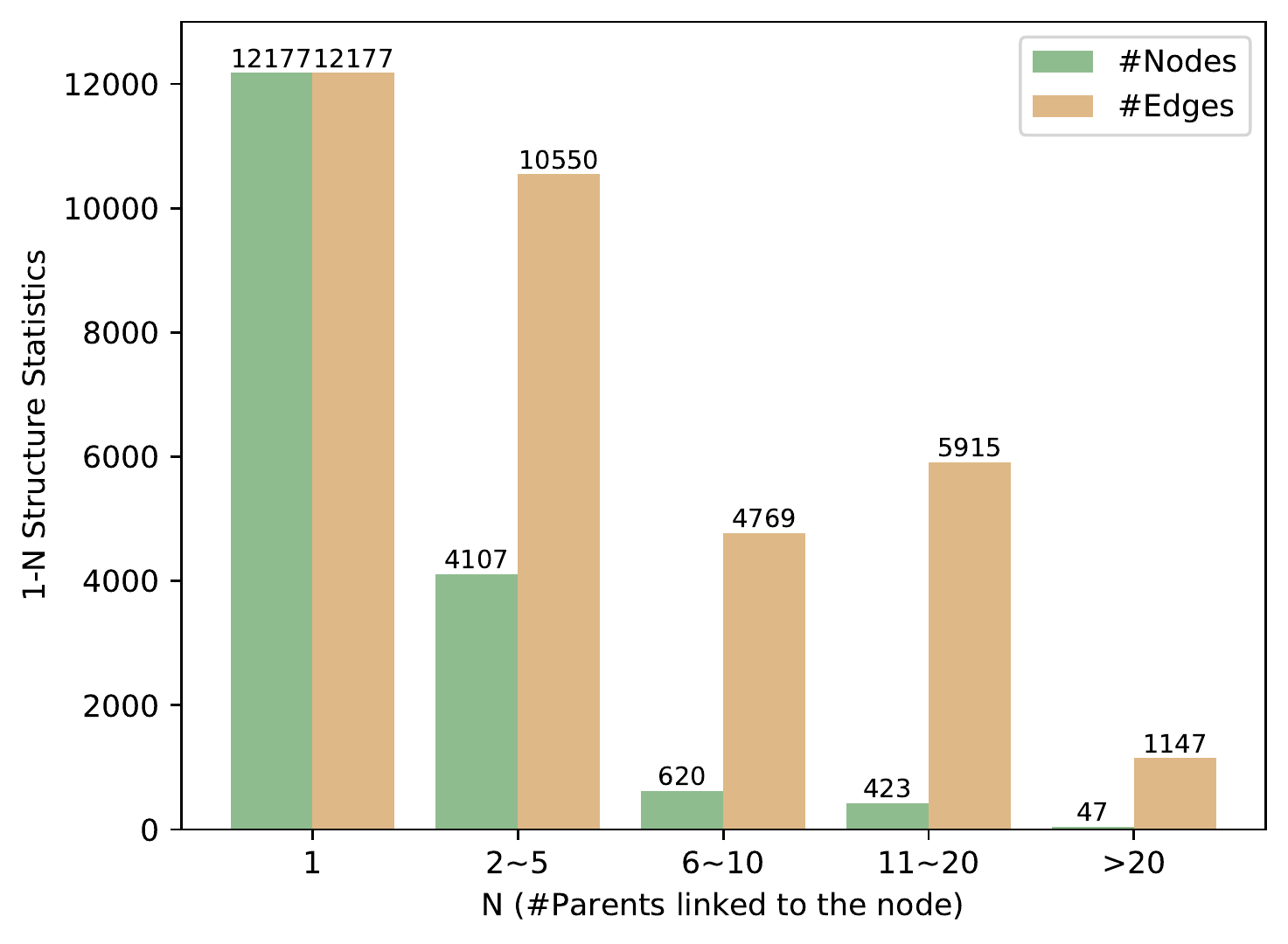}}
\end{subfigure}
\caption{\textbf{Left}: The cloud tree plot of Xiphophorus.
\textbf{Right}: The $1$-$N$ structure statistics in YAGO3-wikiObjects \textit{original} taxonomy. The horizontal axis represents the number of parents linked to a node. The vertical axis counts the number of nodes and edges in the $1$-$N$ structures. For example, there are {$4,107$} nodes that link to $N$ parents for $2\leq N\leq5$, and these links count to {$10,550$} edges.}
\label{fig:more structures}
\Description{The left subfigure is described in the text. The right subfigure shows the following statistics: there are 12,177 nodes that link to 1 parent, and these links count to 12,177 edges. There are 4,107 nodes that link to 2 to 5 parents, and these links count to 10,550 edges. There are 620 nodes that link to 6 to 10 parents, and these links count to 4,769 edges. There are 423 nodes that link to 11 to 20 parents, and these links count to 5,915 edges. There are 47 nodes that link to more than 20 parents, and these links count to 1,147 edges.}
\end{center}
\end{figure}

\subsubsection{Exploring $1$-$N$ Structure}
\label{sec:1-n structure}
A noteworthy difference between the real-world taxonomies and the tree structures is that the taxonomies contain many $1$-$N$ ($1$ child links to multiple parents) cases while in a tree each node except the root is linked to only $1$ parent node. To investigate the advantages of complex hyperbolic geometry on the $1$-$N$ structures, we evaluate the performances of UnitBall, Poincar\'e, and Hyperboloid on predicting the $1$-$N$ edges.
We evaluate on YAGO3-wikiObjects since it contains abundant $1$-$N$ structures. The statistics of the $1$-$N$ structures in YAGO3-wikiObjects \textit{original} taxonomy (\textit{original} means without computing the transitive closure) are given in Figure \ref{fig:more structures}. 

In this experiment, we train the embeddings on the full transitive closure of YAGO3-wikiObjects and then predict the $1$-$N$ edges. The results are reported on Table \ref{tab:more 1-n results} and \ref{tab:1-n results}. We can see that UnitBall has a very small compromise for the $1$-$1$ edges, i.e., the edge pattern of tree structures. Nevertheless, UnitBall outperforms the hyperbolic models largely on $1$-$N$ structures for $N>1$. Furthermore, Unitball has considerably huge improvement for $N>6$, where the hyperbolic embedding models fail to make accurate predictions. Even for nodes that link to more than $20$ parents, UnitBall can have accurate top $10$ predictions for these edges. The results demonstrate that the complex hyperbolic embeddings maintain the advantages in the edge pattern of tree structures as well as handling more complicated hierarchical structures compared with the real hyperbolic embeddings.

\begin{table}[t]
\caption{Results of Hits@1 and Hits@10 on predicting $1$-$N$ edges in YAGO3-wikiObjects. The embedding dimension is 16 for UnitBall while 32 for other models. The best results are shown in boldface.}
\label{tab:more 1-n results}
\begin{center}
\begin{small}
\begin{tabular}{l|rrrrrr}
\toprule
 & \multicolumn{6}{c}{YAGO3-wikiObjects (Hits@1)} \\
  $N$ for $1$-$N$ edges           & $1$ & $2\sim5$ & $6\sim10$ & $11\sim20$ & $>20$ & $>1$ \\
 \midrule
Poincar\'e    & 36.94 & 12.13 & 0.97  & 0.87  & 0.00  & 9.83  \\
Hyperboloid & \textbf{38.24} & 12.20 & 0.97  & 0.95  & 0.00  & 6.95  \\
UnitBall    & 37.64 & \textbf{28.63} & \textbf{18.98} & \textbf{11.82} & \textbf{17.73} & \textbf{26.02}  \\
\bottomrule
    & \multicolumn{6}{c}{YAGO3-wikiObjects (Hits@10)} \\
\midrule
Poincar\'e    &  \textbf{93.10} & 65.48    & 60.11     & 49.09      & 43.97            & 63.29    \\
Hyperboloid & 92.80  & 65.51    & 63.49     & 51.93      & 45.39            & 63.95   \\
UnitBall    & 92.42  & \textbf{76.35}    & \textbf{65.91}     & \textbf{66.19}      & \textbf{70.21}            & \textbf{74.03}  \\
\bottomrule
\end{tabular}
\end{small}
\end{center}
\end{table}

\begin{table}[t]
\caption{Results of MAP on \textbf{predicting $1$-$N$ edges in YAGO3-wikiObjects} and \textbf{reconstructing the leaf node links in Xiphophorus}. The embedding dimension is 16 for UnitBall and 32 for other models. The best results are shown in boldface.}
    \label{tab:1-n results}
    \begin{center}
    \begin{small}
    \begin{tabular}{l|rrrrrr|r}
    \toprule
 & \multicolumn{6}{c}{YAGO3-wikiObjects (MAP)}  & {Xiphophorus (MAP)} \\
$N$ for $1$-$N$ edges     & $1$ & $2\sim5$ & $6\sim10$ & $11\sim20$ & $>20$ & $>1$ & Leaf Links Reconstruction \\
\midrule
Poincar\'e    &  60.73 & 16.87 & 9.28  & 9.31  & 9.46  & 15.29  & 89.75    \\
Hyperboloid & \textbf{61.49} & 15.01 & 9.25  & 9.50  & 9.54  & 13.80  & 89.80         \\
UnitBall    & 58.41 & \textbf{26.73} & \textbf{13.28} & \textbf{10.17} & \textbf{11.64} & \textbf{23.61}   & \textbf{91.95}        \\
\bottomrule
    \end{tabular}
    \end{small}
    \end{center}
\end{table}

\subsubsection{Reconstruction Results on Multitree Structure}

In combinatorics and order-theoretic mathematics, a multitree structure is a directed acyclic graph (DAG) in which the set of vertices reachable from any vertex induces a tree, or a partially ordered set (poset) that does not have four items $a$, $b$, $c$, and $d$ forming a diamond suborder with $a\leq b\leq d$ and $a\leq c\leq d$ but with $b$ and $c$ incomparable to each other (also called a diamond-free poset~\citep{DBLP:journals/jct/GriggsLL12}).\footnote{Here $\leq$ denotes the partial order defined in the graph, e.g., the hypernymy relation.}
Obviously, the multitree structure is not a tree since one child node can have multiple parents in multitree. Note that the multitree structure is also different with $1$-$N$ structure since the multitree has more strict conditions, that is, the multitree is a diamond-free poset. By comparison, the $1$-$N$ structure is more general in taxonomies. For example, the subgraph of YAGO3-wikiObjects \{(\textit{Nei Gaiman}, \textit{is-a}, \textit{British screenwriters}), (\textit{Neil Gaiman}, \textit{is-a}, \textit{British fantasy writers}), (\textit{British screenwriters}, \textit{is-a}, \textit{British writers}), (\textit{British fantasy writers}, \textit{is-a}, \textit{British writers})\} is a $1$-$N$ structure, but it is not a multitree structure, because \textit{British screenwriters} and \textit{British fantasy writers} are incomparable to each other, i.e., there is no partial order between them.

Recall that our synthetic compressed graph is also aggregated from multiple trees, but it is not the multitree either. A compressed graph is aggregated from multiple random trees on the same set of nodes while the trees in a multitree structure share the same leaf nodes. 
Multitrees are widely used to represent multiple overlapping taxonomies over the same ground set.

In this section, we compare the performances of the hyperbolic models and UnitBall on Xiphophorus. Xiphophorus is a multitree dataset that is formed of $160$ mrbayes consensus trees on $26$ Xiphophorus fishes. Its cloud tree plot is in Figure \ref{fig:more structures}.
We reconstruct the edges containing the leaf nodes on Xiphophorus since the leaf links have the practical taxonomic meaning. The results are reported in the last column of Table \ref{tab:1-n results}. The results show that the complex hyperbolic geometry has a stronger ability to represent the multitree structure.

\subsection{Comparison with Trainable Curvature Models}
\label{sec:comparison with trainavle curvature}
Since our work focuses on the representation of single-relation graphs, we do not evaluate the multi-relational knowledge graph embedding models or the neural networks in our main experiments. Nevertheless, to address the concerns of comparison with the trainable curvature models, we compare UnitBall with the hyperbolic knowledge graph embedding method AttH~\citep{DBLP:conf/acl/ChamiWJSRR20} and hyperbolic graph neural networks~\citep{DBLP:conf/nips/ChamiYRL19,DBLP:conf/nips/ZhuP00C020}.

\subsubsection{Comparison with Hyperbolic Knowledge Graph Embeddings}
In this section we evaluate AttH~\citep{DBLP:conf/acl/ChamiWJSRR20} on the single-relation taxonomy link prediction task. We use the released code and tune the hyperparameters on the validation set.\footnote{\url{https://github.com/HazyResearch/KGEmb}.} From the results in Table \ref{tab:AttH results}, we see that UnitBall outperforms AttH in the single hypernymy relation link prediction task. However, UnitBall cannot infer multiple relations like AttH for now. We believe the future work of the complex hyperbolic embeddings will have promising improvements on multi-relational graph embeddings.

\begin{table}[t]
\caption{Evaluation of taxonomy link prediction on YAGO3-wikiObjects (the dimension is $32$ for AttH and $16$ for UnitBall).}
    \label{tab:AttH results}
    \begin{center}
    \begin{small}
    \begin{tabular}{l|rrrr}
    \toprule
            & MAP       & MRR       & Hits@1 & Hits@3     \\
            \midrule
AttH     & 30.22 & 28.47 & 9.10  & 43.83 \\
UnitBall & 33.33 & 31.85 & 15.62 & 47.41   \\
\bottomrule
    \end{tabular}
    \end{small}
    \end{center}
\end{table}

\begin{table}[t]
\caption{Evaluation on link prediction task of GIL paper in ROC AUC (the dimension is $8$ for UnitBall and $16$ for HGCN and GIL).}
    \label{tab:GIL results}
    \begin{center}
    \begin{small}
    \begin{tabular}{l|rrrrr}
    \toprule
            & Disease       & Airport       & Pubmed & Citeseer & Cora     \\
            \midrule
HGCN     & 90.80 & 96.43 & 95.13 & 96.63 & 93.81 \\
GIL & 99.90 & 98.77 & 95.49 & 99.85 & 98.28 \\
UnitBall & 99.09 & 96.61 & 98.80 & 99.34 & 97.64 \\
\bottomrule
    \end{tabular}
    \end{small}
    \end{center}
\end{table}

\subsubsection{Comparison with Hyperbolic GNNs}
\label{app:HGCN}
Although hyperbolic GNNs also involve graph embeddings and can deal with the link prediction task, they make use of not only the edges between nodes but also the node features. The message propagation and attention mechanism make GNNs more flexible to handle various downstream tasks than shallow embeddings.
In this section, we evaluate UnitBall on the link prediction task on the datasets of GIL~\citep{DBLP:conf/nips/ZhuP00C020}. The results of HGCN~\citep{DBLP:conf/nips/ChamiYRL19} and GIL~\citep{DBLP:conf/nips/ZhuP00C020} are copied from Table 2 of the GIL's original paper~\citep{DBLP:conf/nips/ZhuP00C020}. We strictly follow their experimental settings and report the mean results of UnitBall in ROC AUC over 5 running executions.

The results are shown in Table \ref{tab:GIL results}. We can see that UnitBall outperforms HGCN on the five datasets. GIL is slightly better than UnitBall on most datasets while being outperformed by UnitBall on Pubmed. The results are very promising for UnitBall since UnitBall is a shallow embedding approach without deep architecture or feature interaction. We believe the complex hyperbolic embeddings will help to improve the GNNs and bring more insights into geometric deep learning.

\section{Conclusion and Future Work}
\label{sec:conclusion}
In this paper, we present a novel approach for learning the embeddings of hierarchical structures in the unit ball model of the complex hyperbolic space. We characterize the geometrical properties of the complex hyperbolic space and formulate the embedding algorithm in the unit ball model. We exemplify the superiority of our approach over the graph reconstruction task and the link prediction task on both synthetic and real-world data, which cover the various hierarchical structures and two specific structures, namely multitree structure and $1$-$N$ structure. The empirical results show that our approach outperforms the hyperbolic embedding methods in terms of representation capacity and generalization performance. Motivated by our theoretical grounding and empirical success, we believe the complex hyperbolic embeddings will have promising improvements on the knowledge graph embeddings, neural networks, and other related applications.



\bibliographystyle{ACM-Reference-Format}
\bibliography{Reference}


\end{document}